\documentclass[sigconf]{acmart}

\usepackage{graphicx}
\usepackage{multirow}
\usepackage{booktabs}

\usepackage{amssymb}
\usepackage{caption}
\usepackage{subcaption}
\usepackage{float}
\usepackage{pifont}
\usepackage{algorithm}  
\usepackage{algpseudocode} 
\usepackage{amsmath}  
\usepackage{forest}

\newcommand{\cmark}{\ding{51}}  
\newcommand{\xmark}{\ding{55}}  
\AtBeginDocument{%
  }

\setcopyright{acmlicensed}
\copyrightyear{2018}
\acmYear{2018}
\acmDOI{XXXXXXX.XXXXXXX}
\acmConference[Conference acronym 'XX]{Make sure to enter the correct
  conference title from your rights confirmation email}{June 03--05,
  2018}{Woodstock, NY}
\acmISBN{978-1-4503-XXXX-X/2018/06}

\begin{document}
\twocolumn

\newpage

\title{Graph Navier–Stokes Networks}


\author{Zexing Zhao}
\authornote{Work done while the author was an undergraduate student at Northwest A\&F University.  He is currently with Georgia Institute of Technology.}
\affiliation{%
  \institution{Northwest A\&F University}
  \city{Yangling}
  \state{Shaanxi}
  \country{China}}
\email{2019015213zzx@nwafu.edu.cn}

\author{Guangsi Shi}
\authornote{Corresponding author.}
\affiliation{%
  \institution{Corporate Research Center, Midea Group}
  \city{Shanghai}
  \state{}
  \country{China}}
\email{shigs13@midea.com}

\author{Yu Gong}
\affiliation{%
  \institution{Peking University}
  \city{Beijing}
  \country{China}}
\email{2306393391@pku.edu.cn}

\author{Tianyu Wang}
\affiliation{%
  \institution{Fudan University}
  \city{Shanghai}
  \country{China}}
\email{wangty25@m.fudan.edu.cn}

\author{Shirui Pan}
\affiliation{%
  \institution{Griffith University}
  \city{Brisbane}
  \state{Queensland}
  \country{Australia}}
\email{s.pan@griffith.edu.au}

\author{Hongye Cheng}
\affiliation{%
  \institution{Northwest A\&F University}
  \city{Yangling}
  \state{Shaanxi}
  \country{China}}
\email{chy1524727724@nwafu.edu.cn}

\author{Yuxiao Li}
\affiliation{%
  \institution{Bosch}
  \city{Shanghai}
  \country{China}}
\email{yuxiao.li@cn.bosch.com }

\renewcommand{\shortauthors}{Zhao et al.}

\begin{abstract}

Graph Neural Networks (GNNs) have emerged as a cornerstone of deep learning, with most existing methods rooted in graph signal processing and diffusion equations to model message passing. However, these approaches inherently suffer from the oversmoothing problem, where node features become indistinguishable as the network depth increases. Inspired by the Navier-Stokes equations, we introduce \underline{\textbf{G}}raph \underline{\textbf{N}}avier--\underline{\textbf{S}}tokes \underline{\textbf{N}}etworks (GNSN), a novel architecture that transcends conventional diffusion-based message passing by incorporating convection into graph structures. GNSN defines a dynamic velocity field on the graph to govern convection, enabling more efficient and direct message propagation. By adaptively balancing convection and diffusion, GNSN effectively handles datasets with varying levels of homophily. And our analysis reveals, for the first time, a systematic inverse correlation between graph homophily and the strength of convection during message passing. Extensive evaluations across real-world datasets demonstrate that GNSN consistently outperforms state-of-the-art baselines in classification accuracy. Moreover, experimental results further emphasize its effectiveness in alleviating the oversmoothing problem.\footnote{The implementation code supporting this paper is available at: \url{https://github.com/Duckbluee/Graph-Navier-Stokes-Networks}} \footnote{This is an extended preprint version with additional appendix materials. The official conference version appears in KDD 2026. DOI: 10.1145/3770855.3817618.}
\end{abstract}



\received{20 February 2007}
\received[revised]{12 March 2009}
\received[accepted]{5 June 2009}

\maketitle

\section{Introduction}
Fueled by the remarkable ability of graph-structured data to represent entities and their relationships, along with the expressive power of GNNs, research in this field has attracted significant attention. These advancements have been widely applied across various real-world domains, including recommender systems \cite{rs1, rs2}, drug discovery \cite{dd1, dd2}, traffic prediction \cite{tf1, tf2}, and computer vision tasks such as image and semantic segmentation \cite{is}.

In recent years, the integration of GNNs with Neural Ordinary Differential Equations (Neural ODEs) has emerged as a promising research direction \cite{gode, gode2}, extending the discrete message-passing framework of GNNs into the continuous-time domain \cite{cgnn, towards, towards2}. GRAND \cite{grand} introduced the paradigm of "diffusion on graphs", formulated through the discretization of Partial Differential Equations (PDEs), demonstrating that diffusion serves as a natural and effective mechanism for modeling message passing on graphs \cite{diffusiong1, gdif2, pure}. Consequently, research on diffusion-based methods has gained considerable attention, focusing on refining diffusion coefficients and incorporating source or reaction terms within diffusion equations \cite{grand++, gread, acmp, blend}.


However, regardless of the formulation—whether physical diffusion based on Fick's law or graph-based diffusion—the fundamental driving force remains the gradient, which inherently drives the system toward a homogeneous equilibrium \cite{diffusion1, fick}. This mechanism essentially acts as a low-pass filter, and while effective for local smoothing, unconstrained diffusion inevitably leads to the over-smoothing problem \cite{os1, dos}, causing node representations to converge and become indistinguishable. This issue is exacerbated in heterophilic graphs, where indiscriminate smoothing obscures distinct feature distributions across classes \cite{heter}. To address this, the inroduction of reaction or source terms as localized mechanisms has been explored as a way to counteract diffusion by reintroducing unsmoothed features \cite{gread, grand++}. While these approaches successfully mitigate homogenization effects to some extent, they remain constrained within the diffusion equation framework and may not fully resolve the fundamental limitations of diffusion-based message-passing schemes.

Confronted with these bottlenecks, we move beyond the confines of the conventional diffusion process and endeavor to approach the problem from a fresh perspective. Specifically, we conceptualize the graph as a dynamic fluid system, wherein the flux or transmission of scalar quantities within the flow field serves as an analogue to the message flow in the message-passing process \cite{flowx, f2gnn}. Drawing inspiration from the Navier–Stokes equations, we observe that in most physical systems, particularly within engineering contexts, the transport of scalar quantities is predominantly governed by convection rather than diffusion \cite{ns1, ns2}. Convection enables the efficient long-range advection of high-concentration substances with minimal attenuation, thereby preserving steep spatial concentration fronts. In contrast, diffusion operates by smoothing out these fronts through molecular motion, ultimately driving the system toward thermodynamic equilibrium and spatial uniformity \cite{dc, pn}. Motivated by this insight, we contend that message propagation on graphs should not rely solely on diffusion, but instead emerge from the interplay between diffusion and a more direct, dominant convection process. Unlike diffusion, which is driven by feature gradients and inherently promotes homogenization, convection—guided by a learnable velocity field—enables directed information flow that is independent of such gradients. Building on this principle, we propose \underline{\textbf{G}}raph \underline{\textbf{N}}avier–\underline{\textbf{S}}tokes \underline{\textbf{N}}etworks (GNSN), a novel GNN architecture that integrates convection with diffusion, mirroring scalar transport in realistic fluid systems. While diffusion ensures localized feature smoothing, convection facilitates more efficient and directed message passing, leading to a more balanced and robust propagation mechanism. Notably, we introduce a velocity field on graphs to govern convection. By adaptively adjusting the balance between convection and diffusion, GNSN effectively handles graphs with diverse structural properties and varying homophily levels, facilitating localized feature smoothing while mitigating excessive feature homogenization. As shown in Table \ref{tab:table1}, most physics-inspired GNNs for general tasks rely solely on diffusion-based message passing. In contrast, our work introduces a fundamentally new physical paradigm, positioning GNSN as a novel model that transcends diffusion equations by integrating convection and velocity fields inspired by the Navier-Stokes equations. 
This integration extends the conventional paradigm, offering a more comprehensive framework that incorporates both convection- and diffusion-driven message flow on graphs.

\begin{table}[t]
  \centering
  \small
  \caption{A comparison of existing methods.}
  \label{tab:table1}
  \setlength{\tabcolsep}{8pt} 
  \begin{tabular}{lccc}
    \toprule
    Model & Cont. time  & Diff.  & Conv. \\
      & (Neural ODE) & (Gradient) & (Velocity)\\
    \midrule
    FA-GCN\cite{fa-gcn}   & \xmark  & \cmark & \xmark \\
    GPR-GNN\cite{gprgnn}  & \xmark   & \cmark & \xmark \\
    ACM-GCN\cite{acm-gcn}  & \xmark   & \cmark & \xmark \\
    CGNN\cite{cgnn}     & \cmark & \cmark & \xmark \\
    GRAND\cite{grand}    & \cmark & \cmark & \xmark \\
    GRAND++\cite{grand++}  & \cmark & \cmark & \xmark \\
    BLEND\cite{blend}    & \cmark & \cmark & \xmark \\
    ACMP\cite{acmp}     & \cmark & \cmark & \xmark \\
    GREAD\cite{gread}    & \cmark & \cmark & \xmark \\
    \midrule
    Ours     & \cmark & \cmark & \cmark \\
    \bottomrule
    \vspace{-20pt}
  \end{tabular}
\end{table}

During the experimental phase, we evaluate GNSN on thirteen primary datasets, including six heterophilic, six homophilic datasets and a large-scale dataset, with an additional seven datasets detailed in Appendix \ref{appendixB}. Heterophilic datasets are characterized by neighboring nodes that predominantly belong to different classes, whereas homophilic datasets consist of neighboring nodes that typically share the same class. To ensure a robust evaluation, we compare our method against an extensive set of state-of-the-art baselines. Beyond these conventional settings, our study also uncovers an intriguing and previously unexplored regularity in graph transport dynamics. Specifically, we observe for the first time a systematic inverse correlation between graph homophily and the effective strength of convection during message passing--graphs with lower homophily consistently exhibit stronger convection-driven transport. We refer to this phenomenon as the \emph{H-C (Homophily-Convection) Trade-off Law}. In essence, the law suggests that as local neighborhoods become less label-consistent, diffusion-driven smoothing becomes less reliable, and the model correspondingly increases convection intensity to sustain non-local information transport and mitigate excessive homogenization.

The key contributions of our work can be summarized as follows:

\textbf{New Modeling Paradigm:} We propose GNSN, a physics informed framework that transcends purely diffusion-based GNNs by mathematically formalizing convection on graphs, seamlessly integrating both convection and diffusion message flow in message propagation..

\textbf{Theoretical Insight:} We introduce the concept of a velocity field on graphs to govern convection-based message flow, and first time identify a distinct inverse correlation between graph homophily and convection intensity, offering a new perspective on how graph topology influences optimal message-passing dynamics.

\textbf{SOTA Performance:} We perform comprehensive evaluations across a diverse range of datasets, comparing GNSN with several state-of-the-art baselines. Our model consistently demonstrates superior performance, achieving the highest accuracy in the majority of scenarios.

\section{Preliminaries \& Related Work}
In this section, we begin by providing an overview of diffusion process, followed by an introduction to the Navier-Stokes equations and graph homophily. Further related works covering graph message passing, continuous GNNs ODE, graph diffusion methodologies, and graph learning under heterophily are provided in the Appendix \ref{appendixrw}.

\subsection{Diffusion and Graph Smoothing}

In physics, diffusion is a transport phenomenon driven by the thermal motion of molecules. It describes the process by which scalar quantities move from regions of higher concentration or chemical potential to regions of lower concentration or chemical potential, facilitating a relaxation toward equilibrium \cite{diffusion1, diffusion2}.

Let \( c(t) \) be a family of scalar-valued functions representing the distribution of a certain property (e.g., substance concentration) at time \( t \). The specific value of this property at point \( u\) and time \( t \) is denoted by \( c(u, t) \). According to Fick’s first law of diffusion, the flux \( \overrightarrow{J} \) is given by \cite{fick}:
\begin{align}
    \overrightarrow{J} = -D \nabla c, \label{eq:eq1}
\end{align}
where \( \nabla c \) represents the concentration gradient, and \( D \) represents the diffusion coefficient, a parameter that defines the transport properties of the scalar quantities. When adapted to graph domains, this formulation induces a Laplacian local averaging effect, which can be interpreted as a low-pass filtering operation on graph signals. By attenuating high-frequency variations across neighboring nodes, this process leads to more homogeneous node representations in local structure.

\subsection{Navier–Stokes Equations}
The Navier–Stokes equations are a set of partial differential equations that govern the motion of viscous fluids \cite{ns1}. These equations play a fundamental role in various fields, including fluid mechanics \cite{ns2}, meteorology \cite{ns3}, and aerospace engineering \cite{ns4}. Derived from the principles of mass, momentum, and energy conservation, the Navier–Stokes equations provide a rigorous mathematical framework for modeling momentum and scalar transport in fluid systems.

In a fluid system, the Navier–Stokes equations describe momentum changes within fluids, specifically how the velocity field is influenced by pressure, viscosity, and external forces. The fundamental form of the Navier–Stokes equations is given by:
\begin{align}
    \rho\frac{\partial \mathbf{u}}{\partial t}+\rho(\mathbf{u} \cdot \nabla) \mathbf{u}=-\nabla p+\mu \nabla^{2} \mathbf{u}+\mathbf{f}, \label{eq:eq3}
\end{align}
where \( \rho \) is the density of the fluid, \(\mathbf{u} \) is the velocity field of the fluid, \( p\) is the pressure within the fluid, \( \mu \) is the dynamic viscosity of the fluid, \( \nabla^{2} \mathbf{u}\) is the Laplacian of the velocity field, representing the influence of internal viscous forces, and \( \mathbf{f}\) denotes external forces acting on the fluid, such as gravity or electromagnetic forces.

\subsection{Graph Homophily}
The node homophily ratio \( H_{\mathcal{G} }\) of the graph $\mathcal{G} = (\mathcal{V}, \mathcal{E})$ is defined as the average of the homophilic 1-hop neighbor ratio of each node 
\(i\) in \(\mathcal{G}\), and is given by:
\begin{align} H_{\mathcal{G} } = \frac{1}{|\mathcal{V}|} \sum_{i \in \mathcal{V}} h_{\mathcal{N}1}(i) = \frac{1}{|\mathcal{V}|} \sum_{i\in \mathcal{V}} \frac{|{j \in \mathcal{N}_1(i) : y_j = y_i}|}{|\mathcal{N}_1(i)|}, \label{eq:eq5}\end{align}
where \(h_{\mathcal{N}1}(i)\) denotes the homophilic 1-hop neighbor ratio of node \(i\), and \(y_j\) and \(y_i\) are the labels of nodes \(j\) and \(i\), respectively \cite{os1, heter2}.

\section{Proposed Model}
In this section, we present the projection of the Navier–Stokes equations onto graph neural networks (GNNs). We then detail the architectural design of GNSN, followed by its training algorithm. The theoretical motivations, mathematical derivations, and complexity analyses are provided in Appendices \ref{appendixD} and \ref{appendixH}.

\subsection{Navier–Stokes Equation on Graph}
Inspired by the Navier-Stokes equations, we establish an analogy between continuous graphs and physical processes in fluid systems. In GNSN, both convection and diffusion mechanisms are integrated to model message flow. 
\begin{proposition}[Diffusion-Convection Decomposition]
\label{pro:decoupled}
Projecting the transport of scalar quantities in the Navier–Stokes equations onto the message-passing process in GNNs, the message flow at node \(i\) can be decomposed into two fundamental and independent components:
\begin{align}
    \frac{\partial \mathbf{h} _{i}(t)}{\partial t} + \frac{\partial \mathbf{h}^\mathrm{conv} _{i}(t)}{\partial t} =\frac{\partial \mathbf{h}^\mathrm{diff} _{i}(t)}{\partial t} + (S), \label{eq:eq7}
\end{align}

\textbf{Diffusion-driven smoothing}, exclusively driven by the gradient of node features through Laplacian smoothing. 
\begin{align}
\frac{\partial \mathbf{h}^\mathrm{diff} _{i}(t)}{\partial t} = \sum_{j \in \mathcal{N}_{i}} \widetilde{\mathbf{A}}_{i,j} \cdot \nabla \mathbf{h}_{i}, \label{eq:eq9}
\end{align}

\textbf{Convection-driven flow}, driven by the velocity field on the graph, facilitates rapid and direct message propagation between nodes.
\begin{align}
\frac{\partial \mathbf{h}^\mathrm{conv} _{i}(t)}{\partial t} =\mathbf{u}_{i}\sum_{j \in \mathcal{N}_{i}}\mathbf{\widetilde{A}}^\varepsilon _{i,j}\cdot \mathbf{h}_{j}. \label{eq:eq8} 
\end{align}
\end{proposition}
Within the Navier–Stokes equations, the term \(S\) denotes external source or reaction contributions, the detailed modeling of which lies beyond the scope of this discussion. The diffusion component follows prior formulations \cite{grand}. Here, \(\mathbf{u}_i\) represents the node-specific velocity; \(\mathbf{h}_i\) and \(\mathbf{h}_j\) denote the feature vectors of nodes \(i\) and \(j\), respectively; and \(\mathcal{N}_{i}\) indicates the receptive field (i.e., the set of neighboring nodes) of node \(i\). The term \(\nabla \mathbf{h}_i\) corresponds to the gradient of node features. The parameter \(\varepsilon\) controls the sensitivity of \(\widetilde{\mathbf{A}}_{i,j}\) to feature differences. And \(\widetilde{\mathbf{A}}_{i,j}\) denotes the \((i,j)\)-th element of the normalized similarity matrix \(\mathbf{\widetilde{A}} \in \mathbb{R}^{n \times n}\), which shares the same dimensionality as the adjacency matrix \(\mathbf{A}\), and is computed as attention-style normalization:
\begin{align}
\widetilde{\mathbf{A}}_{i, j} := \operatorname{softmax} \left(\frac{(\mathbf{W}_{K} \mathbf{h}_{i})^{\top} (\mathbf{W}_{Q} \mathbf{h}_{j})}{\sqrt{d_{K}}} \right), \label{eq:eq10}
\end{align}
where \(\mathbf{W}_{K}\) and \(\mathbf{W}_{Q}\) are learnable weight matrices for key and query projections, respectively, and \(d_{K}\) is a scaling factor.

\subsection{Convection on Graph}

To ground convection on graph domains that lack continuous spatial coordinates, we adopt a finite-volume-inspired modeling perspective. Rather than relying on explicit spatial derivatives, each node $i$ is treated as a discrete control unit, and its incident edges define fixed interaction channels through which information is exchanged. Under this interpretation, graph message passing can be viewed as the aggregation of feature fluxes across these local units, providing an integral and permutation-invariant description of transport on graphs.

Unlike Euclidean domains where velocity fields are represented by spatial vectors, transport on graphs is inherently constrained by the underlying topology. As a result, the possible flow directions are determined by the adjacency structure, while the remaining degree of freedom lies in the strength of transport along these connections. We therefore decouple the directionality of convection, which is fixed by graph connectivity, from its transport intensity. This leads to the introduction of a virtual graph velocity field $\mathbf{u}_{\mathcal{G}} = (\mathbf{u}_1, \dots, \mathbf{u}_n)$, where each scalar $\mathbf{u}_i$ does not represent a kinematic vector, but instead quantifies the effective intensity of feature transport through node $i$.

\begin{proposition}[Effective Convection Velocity Construction]
\label{pro:velo}
The graph velocity field $\mathbf{u}_{\mathcal{G}}$ consists of node-specific scalar velocities $\mathbf{u}_i$, where $\mathbf{u}_i \in \mathbb{R}^+$ quantifies the effective transport intensity of convection at node $i$. Aligned with flux-based formulations in finite-volume modeling, $\mathbf{u}_i$ is constructed as the time-averaged magnitude of the net incoming feature flux over a node-dependent temporal window $[0, t_i] \subseteq [0, T]$:
\begin{align}
    \mathbf{u}_{i}=\frac{1}{T} \int_{0}^{t_{i}} \left \|F_{i}(\tau)\right \|_2 d \tau, \quad F_{i}(\tau)=\sum_{j \in \mathcal{N}(i)} \alpha_{i,j}(\tau) \mathbf{h}_{j}(\tau) \label{eq:eqvelo}
\end{align}
where $F_i(\tau)$ represents the instantaneous net feature flux entering the control unit of node $i$, and $\alpha_{i,j}(\tau)$ encodes the interaction strength from neighbor $j$  (e.g., edge weights or attention coefficients).
In practice, using a discrete approximation with uniform sampling over $L$ layers (time steps), $\mathbf{u}_i$ is computed as
\begin{align}
        \mathbf{u}_i \approx \frac{1}{L} \sum_{\ell=1}^{L}\left\|\sum_{j \in \mathcal{N}(i)} \alpha_{i,j}^{(\ell)} \mathbf{h}_{j}^{(\ell)} \right \|_2.
    \end{align}
The resulting scalar $\mathbf{u}_i$ serves as a proxy for the overall convection strength at node $i$, summarizing the cumulative intensity of feature transport through the node.
\end{proposition}

Eq.~\eqref{eq:eqvelo} reinterprets the neighbor aggregation as an instantaneous flux accumulation. Consequently, the velocity $\mathbf{u}_i$ is fully determined by the evolving node representations and the interaction weights $\alpha_{i,j}(\tau)$. In the discrete GNN setting, the continuous-time integral naturally reduces to a layer-wise accumulation, where each time step $\ell$ corresponds to the $\ell$-th network layer. Under this view, $\mathbf{u}_i$ captures the aggregated message intensity across layers, providing a node-wise measure of the convection strength throughout the network depth.

Further details on evolution and adaptive integration and numerical stability of the velocity field are provided in Appendix \ref{appendixD}.

\subsection{Position Encoding}
To capture hierarchical graph structures and enrich node representations, we project features from Euclidean space to the Poincaré disk in hyperbolic space as model inputs \cite{pe1, pe3}  (see Appendix \ref{appendixF} for a broader discussion on alternative positional encodings). In Euclidean space, graph convolutions act as low-pass filters, causing oversmoothing and feature homogenization after multiple layers \cite{os1}. In contrast, the negative curvature of hyperbolic space allows features to disperse toward the boundary, mitigating oversmoothing and preserving representational diversity \cite{pe2hgcn}. Specifically, for node features 
\({\mathbf{X}}\) in Euclidean space, their projection 
\(\mathbf{X}^{\prime}\) onto the Poincaré disk model is defined as:
\begin{align}
    \mathbf{X}^{\prime}=\frac{\mathbf{X}}{\|\mathbf{X}\| \cdot\left(1+\sqrt{1+c\|\mathbf{X}\|^{2}}\right)}, \label{eq:eq12}
\end{align}
\begin{align}
    \mathbf{H}=[\mathbf{X} \| \mathbf{X}^{\prime}], \label{eq:eq13}
\end{align}
where \(\|\mathbf{X}\|\) denotes the Euclidean norm of \({\mathbf{X}}\), and \(c\) is the curvature parameter, which controls the degree of hyperbolic geometry's curvature. \(\mathbf{H} \in \mathbb{R} ^{n\times 2d }\) is the new node feature after combination.

\subsection{Overview}
Extending the convection message flow to the entire graph \(\mathcal{G}\) and incorporating the diffusion component and position encoding yields the continuous-time message-passing expression:
\begin{align}
    \frac{\partial \mathbf{H}(t)}{\partial t} + (1-H^{\ast }_\mathcal{G})\left[\mathbf{u}_\mathcal{G}\odot\left(\mathbf{\widetilde{A}}^\varepsilon \mathbf{H}(t)\right) \right]=H^{\ast }_\mathcal{G}\mathbf{\widetilde{L} }\mathbf{H}(t), \label{eq:eq11}
\end{align}
where \(\mathbf{\widetilde{L}} = \mathbf{I} - \mathbf{\widetilde{A}}\) is the normalized graph Laplacian that governs diffusion. The parameter \(\varepsilon\) specifies the order of the adjacency matrix \(\mathbf{A}\), allowing for flexible receptive field control. The scalar \(H^{\ast }_\mathcal{G}\) is a learnable homophily ratio derived from the training data, which adaptively modulate the intensity of convection and diffusion message flows within the message-passing process, enabling adaptability to diverse graph structures. Reliance on either term alone is suboptimal; for example, in heterophilic graphs, excessive diffusion leads to over-smoothing and loss of feature discriminability.

\section{Training Algorithm}
In Algorithm \ref{alg:alg1}, training process optimizes the cross-entropy loss, which is defined as:
\begin{align}
    \mathcal{L} := \sum_{i=1}^n \mathbf{y}_i^T \log \hat{\mathbf{y}}_i, \label{eq:eq14}
\end{align}
where \( \mathbf{y}_i \) is the one-hot encoded ground truth vector for the \( i \)-th training sample, and \( \hat{\mathbf{y}}_i \) represents the predicted probability distribution generated by the model.

\begin{algorithm}[]
\caption{Training Procedure for GNSN}
\label{alg:alg1}
\begin{algorithmic}
\State \textbf{Input:} Processed dataset \( \mathrm{D} \); maximum iterations \( \mathit{max\_iter} \)
\State \textbf{Output:} Optimal model parameters \( \theta^* \)
\vspace{2pt}
\State \textbf{Step 1:} Partition the dataset into training, validation, and test sets: \( \mathrm{D} \to \{\mathrm{D}_{\text{train}}, \mathrm{D}_{\text{valid}}, \mathrm{D}_{\text{test}}\} \)
\State \textbf{Step 2:} Initialize model parameters \( \theta \); set iteration counter \( k \gets 0 \)
\While{$k < \mathit{max\_iter}$}
    \State (a) Sample a mini-batch \( B \) from the training set \( \mathrm{D}_{\text{train}} \)
    \State (b) Perform a forward pass to compute the loss \( \mathcal{L} \) using Equation~(\ref{eq:eq14}) based on the sampled batch
    \State (c) Backpropagate the loss and update the model parameters \( \theta \) using optimizer
    \State (d) Evaluate the updated model on the validation set \( \mathrm{D}_{\text{valid}} \)
    \State (e) If validation performance improves, update and save the best parameters \( \theta^* \)
    \State (f) Increment the iteration counter: \( k \gets k + 1 \)
\EndWhile
\vspace{1pt}
\State \textbf{Step 3:} Return the best-performing parameters \( \theta^* \)

\end{algorithmic}

\end{algorithm}

\section{Experiments}
This section begins with an evaluation of GNSN’s performance on eight public datasets. Ablation and sensitivity studies are then conducted on position encoding, convection-diffusion mechanisms, and velocity field. The section also concludes with an analysis of the model's Dirichlet energy. Additional experimental results and visualization results are provided in the Appendix \ref{appendixA} and \ref{appendixE}, results on supplementary datasets are provided in Appendix \ref{appendixB} and more sensitivity analyses are provided in the Appendix \ref{appendixF}.

\subsection{Node Classification on Real-world Datasets}
\subsubsection{Real-world Datasets}
We evaluate our model against state-of-the-art baselines on a variety of real-world datasets. The main benchmarks include four heterophilic datasets—Texas, Wisconsin, Cornell (from WebKB) \cite{geom} , and Chameleon \cite{squirrel}—and four homophilic datasets: Citeseer \cite{citeseer}, PubMed \cite{pubmed}, Cora \cite{cora}, and Amazon Photo \cite{photo}. Key dataset statistics are provided in Appendix \ref{appendixA}. Additional experiments on homophilic (e.g., CoauthorCS \cite{co}, Amazon Computer \cite{photo}), heterophilic (e.g., Film \cite{film}, Squirrel \cite{squirrel}), and large-scale datasets (e.g., OGB-Arxiv \cite{ogb}) are included in Appendix \ref{appendixB}. All results are averaged over 10 runs with fixed splits following \cite{geom}, reported as mean accuracy with standard deviation.

\subsubsection{Baselines}
To assess the performance of GNSN, we conducted an extensive comparative analysis against a wide range of state-of-the-art baselines across multiple categories. These include models designed for heterophilic graphs (Hetero. oriented), such as Geom-GCN \cite{geom} and PloyGCL \cite{ploy}; methods addressing the oversmoothing issue (Anti. Ovs.), including GCNII \cite{gcnii} and JKNet \cite{jknet}; continuous-time GNN frameworks (Cont.-t.) like CGNN \cite{cgnn}, GDE \cite{gde}, and FROND \cite{frond}; diffusion-based models (Diff. equation) such as GRAND \cite{grand}, GRAFF \cite{graff}, ACMP \cite{acmp}, and GREAD \cite{gread}. Additionally, a more comprehensive comparison encompassing over thirty baseline methods is provided in Appendix \ref{appendixA}.

\subsubsection{Main Results}
Table \ref{tab:main} presents the classification performance across real-world datasets. Our model consistently demonstrates superior performance across eight benchmark datasets, achieving the highest accuracy on the majority of tasks. Notably, it attains state-of-the-art results on Texas, Wisconsin, Chameleon, Cornell, PubMed, Cora and  Photo, underscoring its robustness across both heterophilic and homophilic graph structures. Compared to heterophilic-specialized baselines such as Geom-GCN\cite{geom} and MM-FGCN\cite{fgcn}, GNSN yields substantial improvements, particularly on challenging datasets like Texas, Wisconsin, and Cornell. For homophilic datasets such as PubMed and Cora, GNSN surpasses diffusion-based baselines, including GRAND\cite{grand}, BLEND\cite{blend}, and GCNII\cite{gcnii}, demonstrating its adaptability and effectiveness. By integrating convection and diffusion within the message passing process, GNSN effectively balances feature smoothing with structural diversity preservation, leading to consistently strong performance across diverse graph settings.

\begin{table*}[]
\setlength{\tabcolsep}{3.9pt} 
\renewcommand{\arraystretch}{1.1}
\caption{Results on eight real-world datasets: mean $\pm$ std. dev. accuracy for 10 independent trials. We show the best three results in \textcolor{red}{\textbf{bold}}(first), \textcolor{cyan}{blue}(second), \textcolor{violet}{purple}(third). More results are in Appendix \ref{appendixA}.}
\setlength{\tabcolsep}{6pt} 
\begin{tabular*}{0.99\textwidth}{ccccccccccc}
\toprule
\multicolumn{2}{c}{\textbf{Dataset}} & \textbf{Texas} & \textbf{Wisconsin} & \textbf{Chameleon} & \textbf{Cornell} & \textbf{Citeseer} & \textbf{PubMed} & \textbf{Cora} & \textbf{Photo} \\ 
\multicolumn{2}{c}{Homo. Ratio}     & 0.11 & 0.21 & 0.23 & 0.30 & 0.74 & 0.79 & 0.81 & 0.83 \\ 
\midrule
{Hetero.}    & Geom-GCN\cite{geom}  & 66.76{\tiny$\pm$2.72} & 64.51{\tiny$\pm$3.66} & 60.00{\tiny$\pm$2.81} & 60.54{\tiny$\pm$3.67} & \textcolor{cyan}{78.02{\tiny$\pm$1.15}} & 89.95{\tiny$\pm$0.47} & 85.35{\tiny$\pm$1.57} & 92.35{\tiny$\pm$1.31} \\
{oriented}   & PloyGCL\cite{ploy}     & 88.03{\tiny$\pm$1.80} & 85.50{\tiny$\pm$1.88} & \textcolor{cyan}{71.62{\tiny$\pm$0.96}}   & 82.62{\tiny$\pm$3.11} & \textcolor{red}{\textbf{79.81 {\tiny$\pm$0.85}}} & 87.57{\tiny$\pm$0.62} & 87.15{\tiny$\pm$0.27} & \textcolor{cyan}{93.60{\tiny$\pm$1.37}} \\\midrule
{Anti}  & JKNet\cite{jknet}     & 62.70{\tiny$\pm$8.34} & 53.14{\tiny$\pm$5.22} & 52.63{\tiny$\pm$3.90}  &  59.72{\tiny$\pm$4.60} & 75.99{\tiny$\pm$1.28} & 87.23{\tiny$\pm$0.55} & 86.48{\tiny$\pm$1.04} & / \\
 {Ovs.}      & GCNII\cite{gcnii}     & 77.57{\tiny$\pm$3.83} & 80.39{\tiny$\pm$3.40} & 63.86{\tiny$\pm$3.04} & 77.86{\tiny$\pm$3.79} & 77.33{\tiny$\pm$1.48} & \textcolor{violet}{90.15{\tiny$\pm$0.43}} & \textcolor{violet}{88.37{\tiny$\pm$1.25}} & / \\ \midrule
{}      & GDE\cite{gde}       & 74.05{\tiny$\pm$6.96} & 79.80{\tiny$\pm$5.62} & 47.76{\tiny$\pm$2.08}  & 82.43{\tiny$\pm$7.07} & 76.21{\tiny$\pm$2.11} & 87.80{\tiny$\pm$0.38} & 87.22{\tiny$\pm$1.41} & 92.40{\tiny$\pm$2.00} \\
{Cont.-t.}         & CGNN\cite{cgnn}      & 71.35{\tiny$\pm$4.05} & 74.31{\tiny$\pm$7.26} & 46.89{\tiny$\pm$1.66}  & 66.22{\tiny$\pm$7.69} & 76.91{\tiny$\pm$1.18} & 87.70{\tiny$\pm$0.49} & 87.10{\tiny$\pm$1.35} & 91.30{\tiny$\pm$1.50} \\ 
{}          & FROND\cite{frond}      & 75.56{\tiny$\pm$5.15} & 77.95{\tiny$\pm$6.75} & 71.45{\tiny$\pm$1.98} & 75.36{\tiny$\pm$6.19} & 74.70{\tiny$\pm$1.90} & 79.40{\tiny$\pm$1.50} & 84.80{\tiny$\pm$1.10} & \textcolor{violet}{93.10{\tiny$\pm$1.50}} \\\midrule
{}      & GRAND\cite{grand}     & 75.68{\tiny$\pm$1.25} & 79.41{\tiny$\pm$4.12} & 54.67{\tiny$\pm$2.54} & 82.46{\tiny$\pm$7.09} & 76.46{\tiny$\pm$1.77} & 89.02{\tiny$\pm$0.35} & 87.36{\tiny$\pm$0.96} & 92.30{\tiny$\pm$0.90} \\
{Diff.}       & GRAFF\cite{graff}     & \textcolor{violet}{88.38{\tiny$\pm$3.53}} & \textcolor{violet}{87.45{\tiny$\pm$2.94}} & 71.08{\tiny$\pm$1.75} & 83.24{\tiny$\pm$6.49} & 76.92{\tiny$\pm$1.70} & 88.95{\tiny$\pm$0.52} & 87.61{\tiny$\pm$0.96} & 92.59$\pm$\tiny 0.96 \\
{equation}          & GREAD\cite{gread}     & \textcolor{cyan}{88.92{\tiny$\pm$3.72}} & \textcolor{cyan}{89.41{\tiny$\pm$1.30}} & \textcolor{violet}{71.38{\tiny$\pm$1.31}} & \textcolor{cyan}{87.03{\tiny$\pm$4.95}} & \textcolor{violet}{77.60{\tiny$\pm$1.81}} & \textcolor{cyan}{90.23{\tiny$\pm$0.55}} & \textcolor{cyan}{88.57{\tiny$\pm$0.66}} & 92.65$\pm$\tiny 1.11 \\
{}         & ACMP\cite{acmp}      & 86.20{\tiny$\pm$3.00} & 86.10{\tiny$\pm$4.00} & 70.61{\tiny$\pm$1.56} & \textcolor{violet}{85.40{\tiny$\pm$7.00}} & 75.50{\tiny$\pm$1.00} & 79.40{\tiny$\pm$0.40} & 84.90{\tiny$\pm$0.60} & 91.80{\tiny$\pm$1.10} \\ \midrule
{New}    & GNSN      & \textcolor{red}{\textbf{91.89{\tiny$\pm$2.95}}} & \textcolor{red}{\textbf{90.02{\tiny$\pm$1.94}}} & \textcolor{red}{\textbf{71.85{\tiny$\pm$1.56}}} & \textcolor{red}{\textbf{89.19{\tiny$\pm$5.13}}} & 76.87{\tiny$\pm$1.23} & \textcolor{red}{\textbf{90.41{\tiny$\pm$0.58}}} & \textcolor{red}{\textbf{88.73{\tiny$\pm$0.60}}} & \textcolor{red}{\textbf{95.43{\tiny$\pm$1.26}}} \\
\bottomrule
\label{tab:main}
\vspace{-15pt}
\end{tabular*}
\end{table*}

\textbf{Large-scale Evaluation}
In addition to the eight real-world node classification benchmarks, we further evaluate the performance of GNSN on the large-scale OGB-Arxiv \cite{ogb} dataset to examine its scalability and effectiveness under realistic graph sizes. And we compare GNSN with a set of representative baseline models, including GCN~\cite{gcn}, GAT~\cite{gat}, CGNN~\cite{cgnn}, GDE~\cite{gde}, GRAND~\cite{grand}, and BLEND~\cite{blend}. The experimental results are reported in Tab. \ref{tab:ogb_arxiv_results}. As shown in the results, GNSN maintains strong and stable performance even on this large-scale dataset, demonstrating its ability to effectively handle long-range information propagation in large graphs. Notably, GNSN consistently outperforms diffusion-based baselines such as GRAND and BLEND, and achieves performance competitive with GAT, indicating that GNSN remains effective beyond small and medium-sized graphs.

\begin{table*}[h]
\small
\centering
\caption{Results on large-scale dataset: mean $\pm$ std. dev. accuracy for 10 independent trials. We show the best three results in \textcolor{red}{\textbf{bold}}(first), \textcolor{cyan}{blue}(second), \textcolor{violet}{purple}(third).}
\begin{tabular}{cccccccc}
\toprule
\textbf{Dataset} & \textbf{GCN}\cite{gcn} & \textbf{GAT}\cite{gat} & \textbf{CGNN}\cite{cgnn} & \textbf{GDE}\cite{gde} & \textbf{GRAND}\cite{grand} & \textbf{BLEND}\cite{blend} & \textbf{GNSN} \\
\midrule
OGB-Arxiv & 71.74\tiny$\pm$0.29 & \textcolor{red}{\textbf{73.01\tiny$\pm$0.19}} & 58.70\tiny$\pm$2.50 & 56.66\tiny$\pm$10.9 & 72.23\tiny$\pm$0.20 & \textcolor{violet}{72.56\tiny$\pm$0.10} & \textcolor{cyan}{72.92\tiny$\pm$0.51} \\
\bottomrule
\end{tabular}
\label{tab:ogb_arxiv_results}
\end{table*}

\subsubsection{H-C Trade-off Law}

In GNSN, we introduce a graph velocity field to govern convection-driven message flow. Each node is associated with an independent velocity, collectively forming a node-wise velocity field. Rather than being a fixed parameter, the velocity field is dynamically induced by the evolving feature distribution and local graph structure.

\begin{figure}[htbp]
  \centering
  \includegraphics[width=0.45\textwidth]{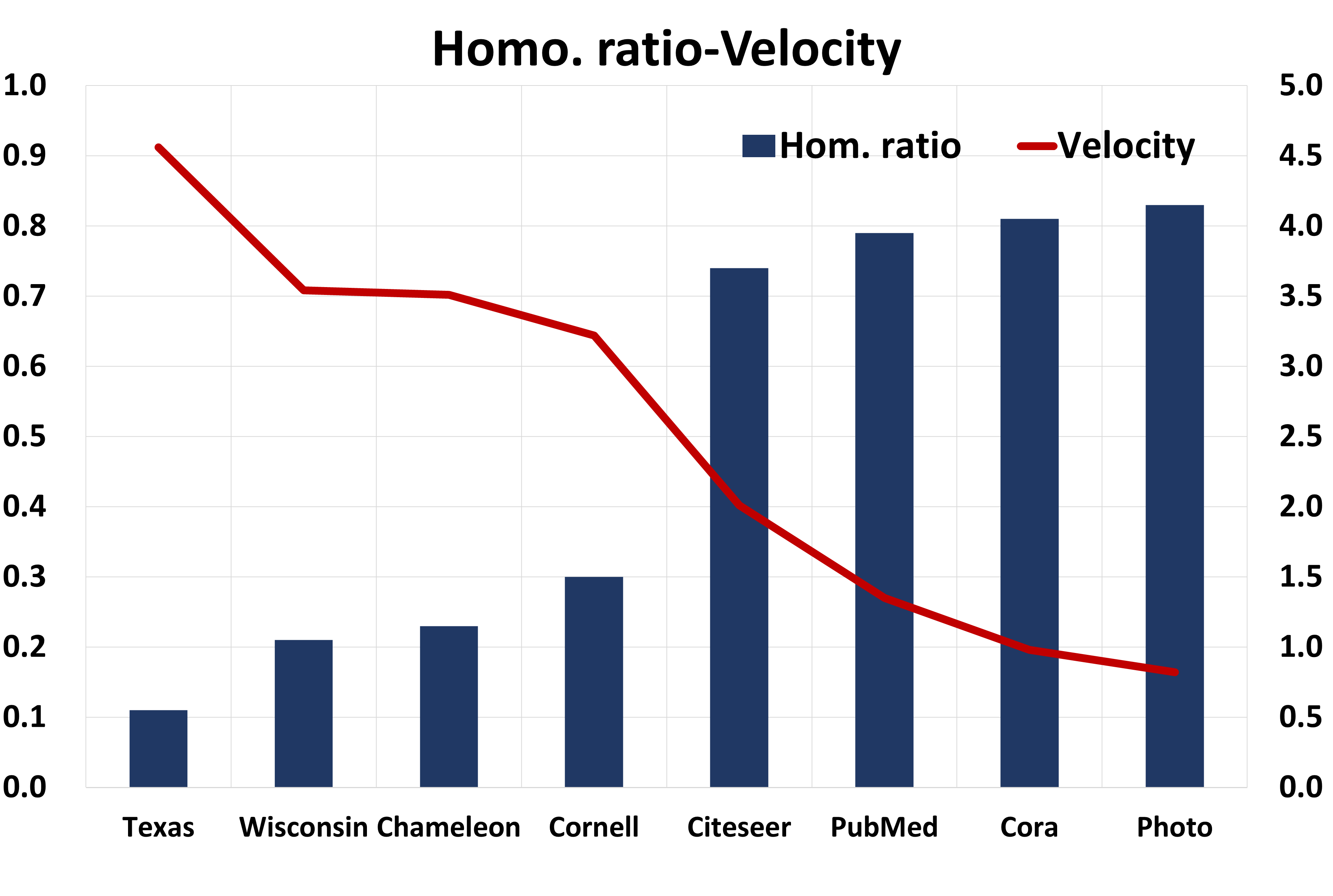}
  \Description{Average velocity magnitude across datasets.}
  \caption{Average velocity magnitude across datasets.}
  \label{fig:fig1}
\end{figure}

To provide an intuitive understanding of the dominant role of convection on graph message passing, we compute the average node velocity magnitude for each dataset, as shown in Figure~\ref{fig:fig1}.

A clear inverse correlation is observed between the average velocity magnitude and the dataset homophily level: heterophilic datasets (e.g., Texas, Wisconsin, Chameleon, Cornell) consistently exhibit higher average velocities, whereas homophilic datasets (e.g., Citeseer, PubMed, Cora, Photo) display lower values. This empirical pattern reveals a systematic trade-off between homophily ratio and convection dominance, which we refer to as the \emph{H-C Trade-off Law}. Intuitively, in heterophilic graphs, local diffusion is prone to excessive feature homogenization, and nodes that facilitate cross-region information exchange develop stronger convection intensity to sustain non-local message transport and mitigate over-smoothing. Conversely, in homophilic graphs, where local neighborhoods are more semantically consistent, diffusion dominates the propagation process, resulting in a reduced reliance on convection.

To further investigate the microscopic structure of the velocity field, we visualize node velocities on subgraphs extracted from the heterophilic Texas dataset and the homophilic Cora dataset.

\begin{figure}[htbp]
  \centering
  \includegraphics[width=0.45\textwidth]{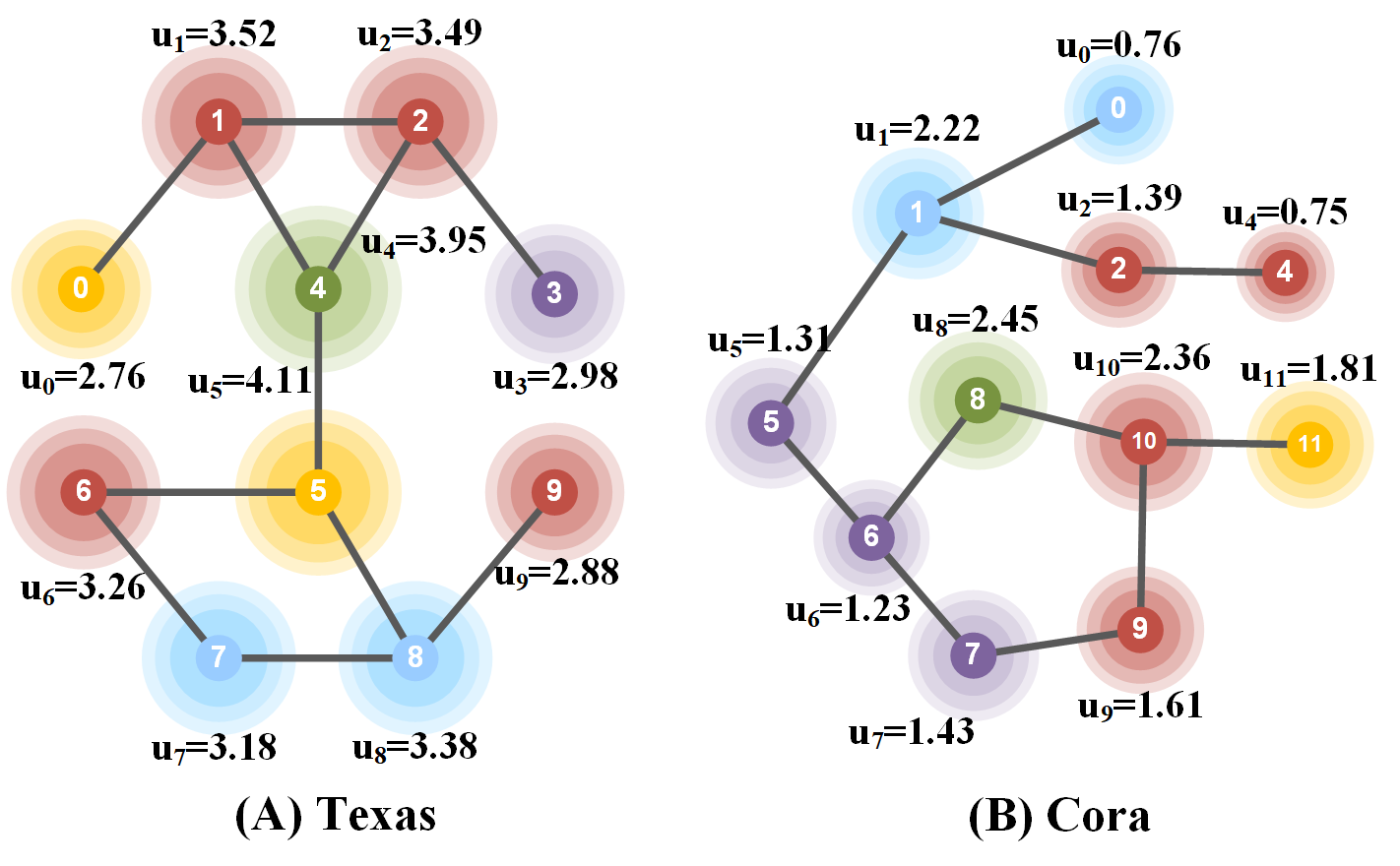}
  \Description{Visualization of the velocity of nodes on a subgraph from the Texas and Cora datasets. (The size of the colored circles represents the magnitude of the velocity. The subgraph is selected from non-central nodes in the datasets.)}
  \caption{Visualization of the velocity of nodes on a subgraph from the Texas and Cora datasets. (The size of the colored circles represents the magnitude of the velocity. The subgraph is selected from non-central nodes in the datasets.)}
  \label{fig:fig2}
  \vspace{-10pt}
\end{figure}

As illustrated in Figure~\ref{fig:fig2}, nodes in the Texas subgraph exhibit systematically higher velocity magnitudes than those in the Cora subgraph, consistent with the macroscopic trend. At the local level, node velocity is strongly coupled with neighborhood composition.
Nodes with fewer same-class neighbors (i.e., lower local label agreement) tend to develop larger velocities, while nodes embedded in more homogeneous neighborhoods exhibit weaker convection intensity.
Moreover, nodes with similar local structural patterns consistently present comparable velocity magnitudes across both datasets.
In homophilic graphs, although this structural correspondence persists, the overall velocity scale remains low, reflecting the dominant role of diffusion and the diminished contribution of convection.

\subsection{Ablation Studies \& Sensitivity Studies}

\begin{figure}[htbp]
  \centering
  \includegraphics[width=0.45\textwidth]{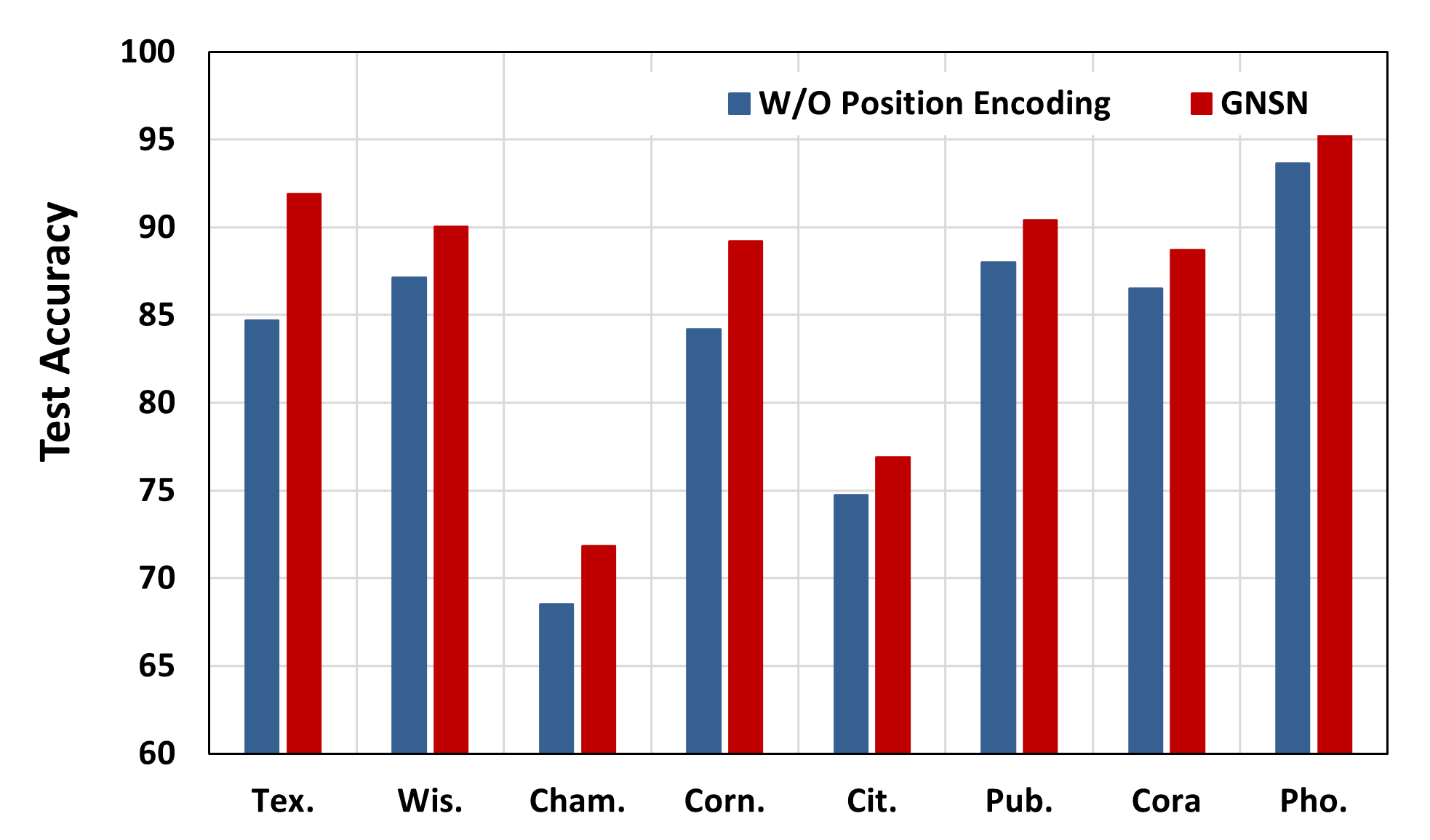}
  \Description{Ablation study on position encoding in hyperbolic space.}
  \caption{Ablation study on position encoding in hyperbolic space.}
  \label{fig:fig3}
\end{figure}

\textbf{Position Encoding:} 
We conducted an ablation study to assess the importance of position encoding in GNSN. As illustrated in the Figure \ref{fig:fig3}, we compared the performance of the full GNSN model with a variant that omits the hyperbolic position encoding across various datasets. The results clearly show that removing position encoding leads to a consistent drop in performance across all datasets, with particularly significant degradation observed on heterophilic datasets. This underscores the essential role of hyperbolic position encoding in improving the model’s ability to capture structural information and address the challenges posed by complex data distributions.

\begin{figure}[htbp]
  \centering
  \includegraphics[width=0.45\textwidth]{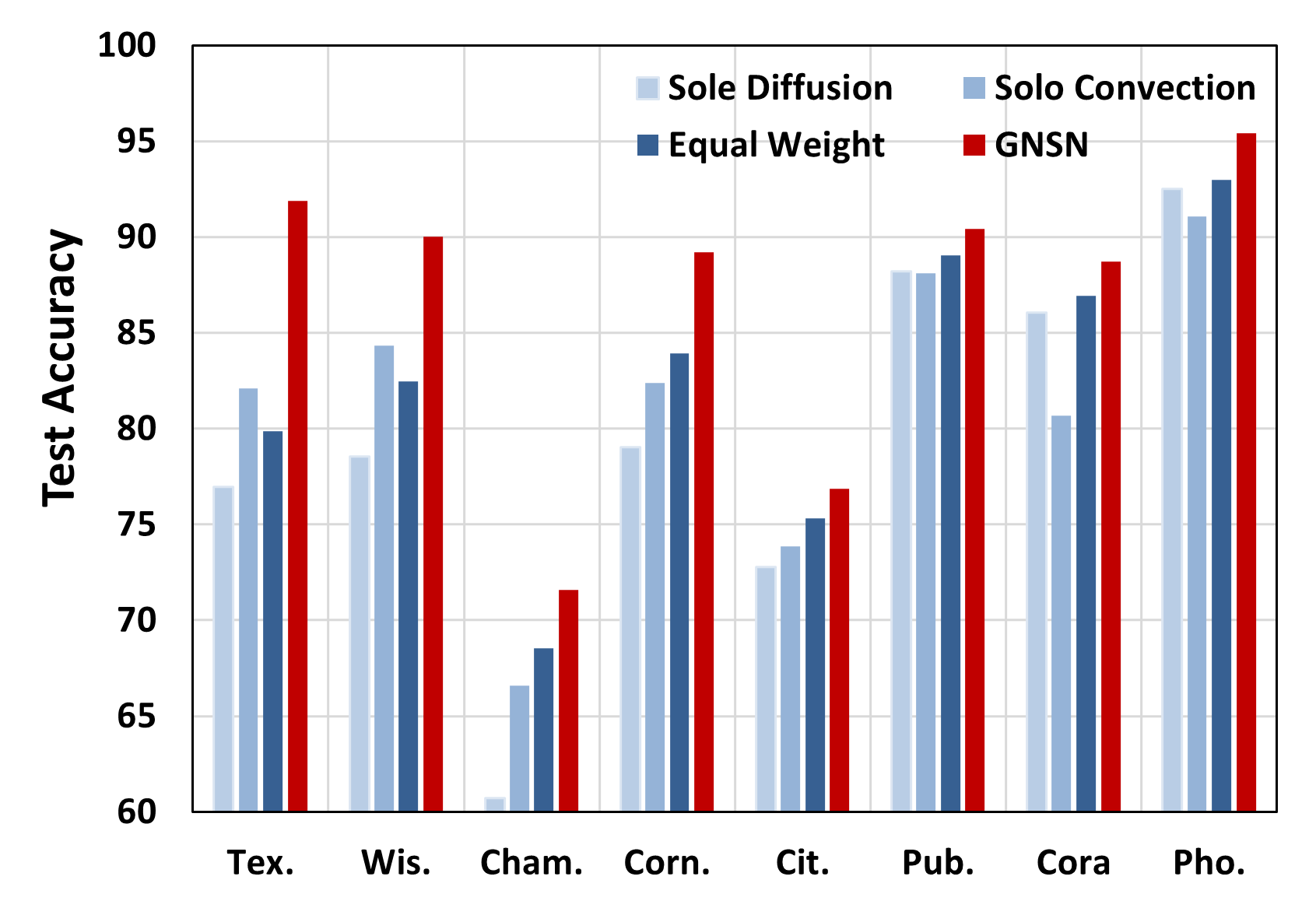}
  \Description{Ablation study on convection and diffusion in message passing.}
  \caption{Ablation study on convection and diffusion in message passing.}
  \label{fig:fig4}
  \vspace{-10pt}
\end{figure}

\textbf{Convection\&Diffusion:} 
To further examine the roles of convection and diffusion across datasets with varying homophily levels, we designed three experimental baselines for comparison with GNSN: one applying only diffusion, one applying only convection, and one assigning equal weights to both. As shown in Figure \ref{fig:fig4}, none of these alternatives outperform GNSN, which dynamically adjusts the balance between convection and diffusion based on the dataset’s homophily level. The performance gap is particularly evident on heterophilic datasets. Notably, on datasets such as Texas, Wisconsin, Cornell, and Citeseer, models utilizing only convection surpass those relying solely on diffusion, further underscoring the crucial role of convection in effective message passing under heterophilic conditions.

\begin{figure}[htbp]
  \centering
  \includegraphics[width=0.45\textwidth]{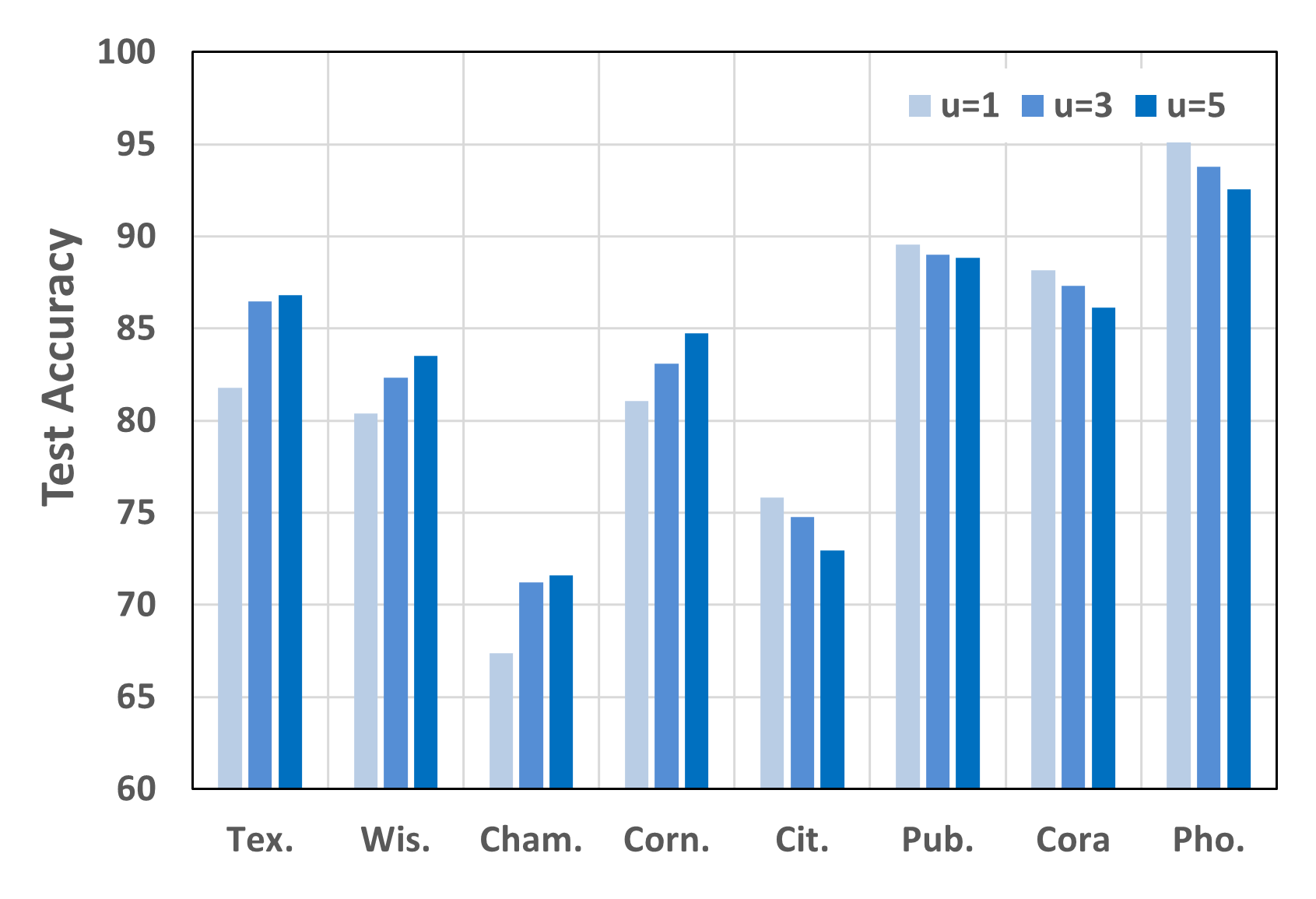}
  \Description{Sensitivity study on the average velocity magnitude.}
  \caption{Sensitivity study on the average velocity magnitude.}
  \label{fig:fig5}
  \vspace{-5pt}
\end{figure}

\textbf{Velocity Field:} To further investigate the impact of different velocity fields on GNSN, we conducted experiments in which all node velocities were unified and fixed as immutable parameters. Three distinct velocity gradients were tested to assess GNSN’s performance across various datasets under different velocity settings. As shown in Figure \ref{fig:fig5}, the benefits of increasing velocity diminish as the homophily ratio increases, consistent with the patterns observed in Figure \ref{fig:fig1}. On heterophilic datasets, where convection is the dominant mechanism, higher velocities—corresponding to stronger convection—yield substantial performance improvements. Conversely, on homophilic datasets, where diffusion naturally plays a more significant role, excessive velocity disrupts the balance between convection and diffusion, leading to performance degradation. These trends are further corroborated by the results in Figure \ref{fig:fig4}. Additional experimental results are provided in Appendix \ref{appendixF}.

\subsection{Oversmoothing \& Dirichlet Energy}

\subsubsection{The Dirichlet Energy}
The degree of oversmoothing can be analyzed from the perspective of Dirichlet energy \cite{os1, DE1}, as defined in \cite{DE3}. The Dirichlet energy \( E\left(\mathbf{H}, \mathbf{A}\right) \) for the node hidden features \( \mathbf{H} \) of an undirected graph \( \mathcal{G} \) is expressed as follows:
\begin{align}
E\left(\mathbf{H}, \mathbf{A}\right) & = \frac{1}{N} \sum_{i \in \mathcal{V}} \sum_{i \in \mathcal{N} .} \mathbf{A}_{[i, j]}\left\|\mathbf{h}_{i}-\mathbf{h}_{j}\right\|^{2}, \label{eq:eq15}
\end{align}
where \( \mathbf{h}_i \) and \( \mathbf{h}_j \) represent the \(i\)-th and \(j\)-th rows of \( \mathbf{H} \), respectively. The oversmoothing phenomenon refers to the tendency of all node features to converge to constant values as the depth of the GNNS increases. Consequently, \( E \) asymptotically decays to zero over time. 

\subsubsection{Experimental Configuration}
To assess the efficacy of GNSN in mitigating oversmoothing, we utilize a synthetic dataset, cSBMs \cite{csbms}, which consists of an undirected graph with 100 nodes positioned in a two-dimensional space and divided into two classes. Edges are randomly generated between nodes with a connection probability of \(p=0.9\). We examine the layer-wise Dirichlet energy of a 30-layer GNN and perform an in-depth comparison across various categories of GNN methods.

\begin{figure}[htbp]
  \centering
  \includegraphics[width=0.45\textwidth]{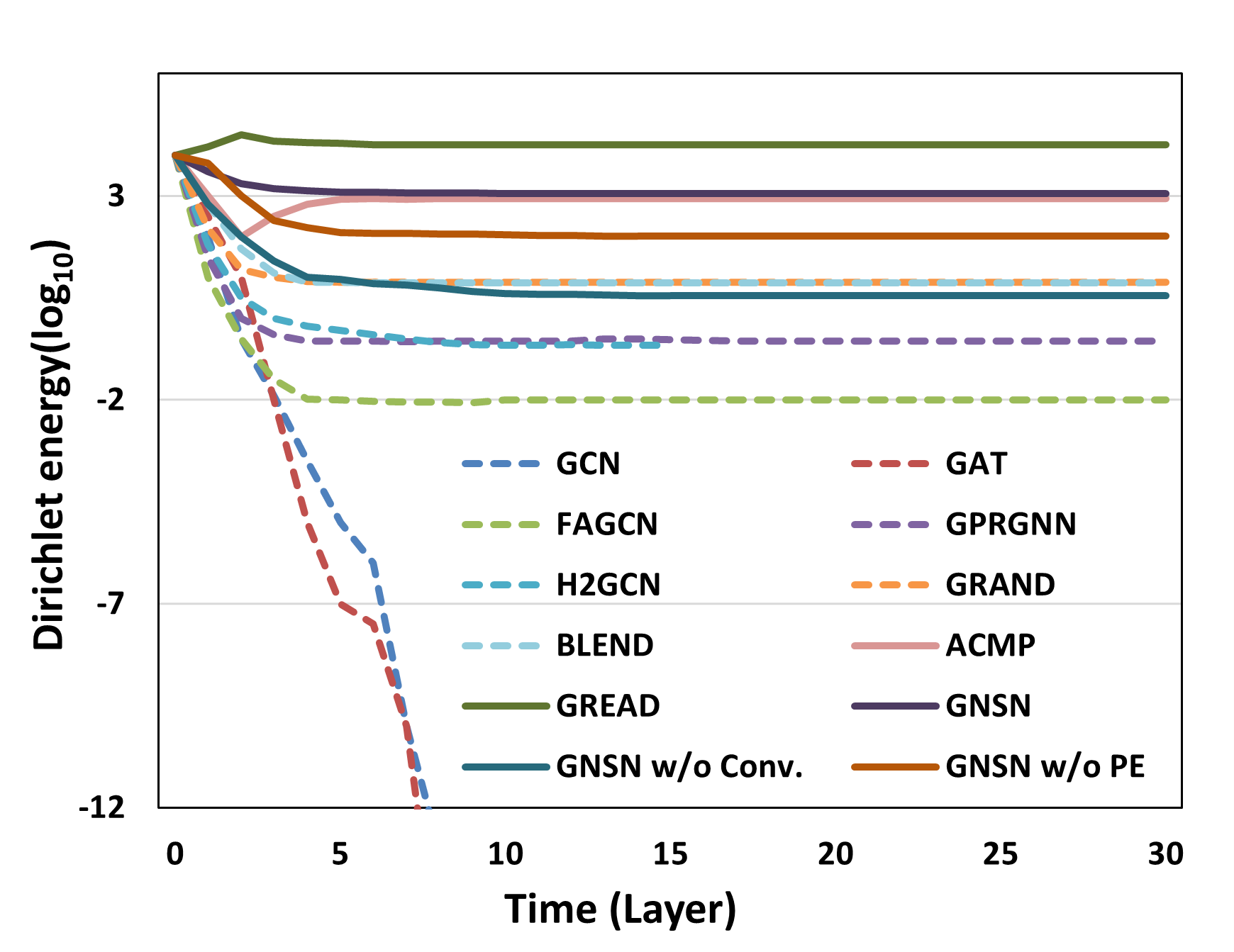}
  \Description{Evolution of the Dirichlet energy on the synthetic random graph.}
  \caption{Evolution of the Dirichlet energy on the synthetic random graph.}
  \label{fig:fig6}
\end{figure}

\subsubsection{Experimental Results}
As illustrated in Figure \ref{fig:fig6}, traditional GNNs, such as GCN and GAT, exhibit a pronounced tendency toward oversmoothing, evidenced by the rapid exponential decay of Dirichlet energy to near zero within just a few layers. This behavior signifies the homogenization of node features, wherein they converge to nearly identical values across the graph. While GRAND incorporates a diffusion term with learnable diffusion coefficient, which delays the onset of oversmoothing, it does not completely address the underlying issue. In contrast, GNSN demonstrates a remarkable resilience to oversmoothing, effectively preserving feature diversity while achieving a balance with smoothing processes. More analyses are provided in Appendix \ref{appendixD}.

\section{Conclusion}

Drawing inspiration from the Navier-Stokes equations, we depart from traditional GNN frameworks that rely solely on diffusion-based mechanisms and introduce GNSN, a novel paradigm that integrates both convection and diffusion into the message-passing process. By formulating a dynamic velocity field on graphs, GNSN enables adaptive convection-driven transport, allowing information to propagate more efficiently. Extensive evaluations across real-world datasets demonstrate that GNSN consistently outperforms state-of-the-art baselines, achieving superior classification accuracy and robust performance under varying homophily conditions. 
Experiments on synthetic datasets further confirm GNSN's ability to mitigate over-smoothing, enabling the construction of deeper and more expressive GNN architectures. Beyond empirical performance gains, we observe an inverse correlation between homophily and the effective strength of convection, which we summarize as the \emph{H-C Trade-off Law}. This observation provides a complementary perspective for understanding how diffusion and convection jointly shape information propagation on graphs.

\begin{acks}
To Robert, for the bagels and explaining CMYK and color spaces.
\end{acks}

\newpage
\balance
\bibliographystyle{ACM-Reference-Format}
\bibliography{reference}

\newpage
\appendix
\onecolumn

\section{Additional Related Works}
\label{appendixrw}
\subsection{Graph Massage Passing}

Message passing on graphs is a fundamental mechanism for feature updates in GNNs \cite{mp, acmp}. Given an undirected graph $\mathcal{G} = (\mathcal{V}, \mathcal{E})$, where $\mathcal{V}$ and $\mathcal{E}$ represent the sets of nodes and edges, respectively, let $\mathbf{x}_{i}^{k}$ denote the features of node $i$ at layer $k$ and $\mathbf{A}_{i,j}$ represent the adjacency relation between node $i$ and node $j$. The feature update rule is given by:
\begin{align} \mathbf{x}_i^{k+1} = \epsilon^{k+1}\left(\mathbf{x}_{i}^{k}, \Box_{j \in \mathcal{N}_1(i)} \phi^{k+1}\left(\mathbf{x}_i^{k}, \mathbf{x}_{j}^{k}, \mathbf{A}_{i,j}\right)\right), \label{eq:eq4}\end{align}

where $\Box$ denotes a differentiable, permutation-invariant aggregater (e.g. sum, mean, or max). $\epsilon$ and $\phi$ are differentiable functions, typically implemented as MLPs, and $\mathcal{N}_1(i)$ represents the set of one-hop neighbors of node $i$. Unlike traditional deep learning architectures, where depth primarily determines hierarchical feature extraction, in GNNs, the number of layers controls the range of message passing, determining how many orders of neighboring nodes contribute to feature aggregation and thereby governing the receptive field of each node.

\subsection{Spectral Graph Neural Networks}

Spectral GNNs define graph convolution as filtering in the Laplacian eigenspectrum. ChebNet\cite{chebnet} and GCN\cite{gcn} established polynomial approximation as the standard approach, while APPNP\cite{appnp} decoupled feature transformation from propagation via Personalized PageRank. GPR-GNN\cite{gprgnn} extended this by adaptively learning generalized PageRank weights to jointly optimize feature and topological information extraction across both homophilic and heterophilic graphs. This motivated a series of more expressive polynomial filter designs: BernNet\cite{bernnet} leveraged Bernstein polynomial bases with non-negative coefficient constraints, ChebNetII\cite{cheb2} addressed overfitting through Chebyshev interpolation, JacobiConv\cite{jacob} adopted orthogonal Jacobi bases, and OptBasisGNN\cite{opt} learned data-driven optimal bases. UniFilter\cite{uni} further addressed the limitation of fixed polynomial bases by theoretically analyzing the relationship between spectral basis properties and graph heterophily, constructing an adaptive universal basis that simultaneously mitigates over-smoothing and over-squashing. TFE-GNN\cite{tfe} proposed a triple filter ensemble that adaptively extracts both homophilic and heterophilic information without relying on polynomial computations or coefficient constraints. In self-supervised learning, PolyGCL\cite{ploy} introduced spectral polynomial filters into graph contrastive learning, constructing low-pass and high-pass views to overcome the inherent homophily bias of conventional GCL encoders.

\subsection{Continuous GNNs via Ordinary Differential Equations}

Traditional graph neural networks are typically formulated as discrete, layer-wise
message passing architectures, where node representations are updated iteratively
across a finite number of layers. This discrete formulation can be interpreted as
a forward Euler discretization of an underlying continuous dynamical system,
a connection that has been widely discussed in the context of residual networks
and neural ordinary differential equations (Neural ODEs) \cite{ODEs}.

Concretely, let $\mathbf{X}^{(k)}$ denote node features at the layer $k$.
A residual-style update of the form
\begin{align}
    \mathbf{X}^{(k+1)} = \mathbf{X}^{(k)} + f(\mathbf{X}^{(k)}, \mathcal{G})
\end{align}
can be viewed as a single Euler step approximating the continuous-time evolution
\begin{align}
    \frac{d\mathbf{X}(t)}{dt} = f(\mathbf{X}(t), \mathcal{G}),
\end{align}
where $t$ denotes a continuous depth or time step.
This observation motivates modeling graph representation learning as the solution
of an ODE defined on graph-structured data.

Building on Neural ODEs \cite{ODEs}, several works have extended the
continuous-time formulation to graph domains.
In Continuous Graph Neural Networks (CGNNs \cite{cgnn}), where node embeddings evolve according to an ODE whose dynamics
are governed by graph-based operators.
Similarly, Graph Differential Equations (GDEs \cite{gde}) provided a general framework for defining continuous-time message passing
dynamics on graphs and analyzing their stability properties.

In this continuous-depth setting, the evolution of node representations is no longer tied to a fixed number of layers, but is instead determined by integrating the ODE over a time interval. Numerical ODE solvers, such as Euler or Runge-Kutta methods, are employed to approximate the solution, introducing an explicit connection between model behavior, numerical discretization, and computational cost. This perspective provides a unified framework within which different graph neural
architectures can be understood as specifying different forms of the dynamical function $f(\cdot)$, while sharing a common continuous-time foundation.

\subsection{Graph Diffusion Methodologies}

Diffusion-based models offer a continuous, mathematically grounded alternative to discrete message-passing schemes by interpreting node feature evolution as a dynamic transport process. While early approaches relied on fixed Laplacian operators that mimic isotropic heat diffusion \cite{gdif1}, recent advancements interpret graph representation learning as the integration of continuous dynamical systems, often modeled as Ordinary Differential Equations (ODEs) \cite{diffusiong1, diffusiong2}. A foundational framework in this domain is GRAND \cite{grand}, which formulates graph diffusion as:
\begin{align}
    \frac{\partial \mathbf{X}(t)}{\partial t} = \mathrm{div}\!\left[\mathbf{G}(\mathbf{X}(t), t)\nabla \mathbf{X}(t)\right],
    \label{eq:general_diffusion}
\end{align}
where $\nabla$ and $\mathrm{div}$ are graph gradient and divergence operators, and $\mathbf{G}$ is a learnable diffusivity tensor that modulates information flow based on feature similarity. This continuous-depth formulation generalizes discrete aggregation but introduces challenges such as over-smoothing, where node representations converge to indistinguishability over long integration times.

Subsequent research can be viewed as a systematic exploration of the ODE in Eq. \eqref{eq:general_diffusion} to address these limitations. 

\textbf{Reaction and Source Terms.} To mitigate over-smoothing, several methods augment the diffusion equation with non-conservative terms. GRAND++ \cite{grand++} introduces a source term to continuously inject initial features, preventing representation collapse. Similarly, reaction-diffusion models like GREAD \cite{gread} employ reaction terms to regulate feature evolution, explicitly constructing boundaries between distinct node classes.

\textbf{Geometric Enrichment.} Other works reconsider the geometric priors of the diffusion operator. Non-Euclidean approaches, such as BLEND \cite{blend} and cellular sheaf diffusion \cite{sheaf}, enrich the underlying graph geometry. These methods decouple feature evolution from the rigid graph topology, improving expressiveness in heterophilic settings.

\textbf{Convection and Transport.} Extending beyond symmetric smoothing, recent models incorporate directional transport effects. CDE \cite{cde} introduces convection terms to model asymmetric information flow. However, these convection terms are typically derived from local feature gradients, meaning the transport dynamics remain tightly coupled to the very feature variations they aim to modulate.

Collectively, these advancements highlight the critical role of the ODE structure in shaping inductive biases. Existing methods predominantly focus on stabilizing diffusion or enriching geometry, yet the coupling between transport dynamics and local features remains a constraint for modeling complex propagation patterns.

\subsection{Graph Learning under Heterophily}

In addition to the node-level homophily ratio $\mathcal{H_{G}}$ discussed in the main text, a widely used metric for quantifying label homophily is the edge homophily ratio $\mathcal{H}_{edge}$, defined as
\begin{align}
\mathcal{H}_{edge}
= \frac{|\{(i,j)\in \mathcal{E}:\; y_i = y_j\}|}{|\mathcal{E}|},
\label{eq:edge-homophily}
\end{align}

which measures the fraction of intra-class edges in the graph \cite{h2gcn}.
While intuitive, recent studies have shown that $\mathcal{H}_{edge}$ can be misleading in the presence of strong class imbalance or varying numbers of classes. In particular, some commonly used heterophilic benchmarks have been found to contain artifacts such as duplicate nodes or information leakage, calling into question the reliability of reported performance gains \cite{critical}.

To address these issues, Platonov et al. \cite{critical} proposed an
adjusted homophily ratio $\mathcal{H}_{adj}$ measure that corrects the observed intra-class connectivity by its expected value under a degree-preserving null model:
\begin{align}
\mathcal{H}_{adj}=\frac{
\mathcal{H}_{edge} - \sum_{k=1}^{C}\frac{D_k^2}{(2|\mathcal{E}|)^2}
}{1 - \sum_{k=1}^{C}\frac{D_k^2}{(2|\mathcal{E}|)^2}},
\quad
D_k := \sum_{i:\, y_i=k} d(i),
\label{eq:adjusted-homophily}
\end{align}

where $C$ denotes the number of classes and $d(i)$ is the degree of node $i$.
This adjusted metric enables more reliable comparisons across datasets with different class distributions and graph densities.

Beyond evaluation metrics, a substantial body of work has proposed GNN architectures explicitly designed for heterophilic graphs, where naive
one-hop aggregation may mix incompatible signals.

\textbf{Multi-hop and propagation redesign:}
A class of methods addresses heterophily by expanding aggregation beyond immediate
neighbors or learning flexible propagation operators.
Representative approaches, including MixHop\cite{mixhop}, GPR-GNN\cite{gprgnn}, and FA-GCN\cite{fa-gcn}, combine messages
from multiple hops or propagation scales, thereby reducing the dominance of
heterophilic one-hop edges and enabling information integration over broader
structural contexts. More recently, adaptive-depth methods~\cite{adaptive}
extend this by assigning node-specific aggregation depths guided by local homophily
and neighborhood structure, with theoretical guarantees on the optimality of
node-wise propagation range.

\textbf{Geometry- and topology-aware neighborhood construction:}
Another line of work incorporates geometric or topological priors to redefine node neighborhoods independently of feature similarity.
Models such as Geom-GCN\cite{geom} construct latent neighborhoods based on structural or geometric relations, suggesting that heterophily in label space does not necessarily preclude meaningful organization in the underlying graph topology.

\textbf{Compatibility-aware aggregation:}
Complementary to the above, diagnostic and design studies such as H2GCN
\cite{h2gcn} demonstrate that many standard GNNs fail under low homophily and identify effective architectural principles, including ego-neighbor embedding separation, the use of higher-order neighborhoods, and the combination of intermediate representations. Architectures following this paradigm explicitly decouple self-information from neighbor messages, preserving discriminative features under heterophilic mixing.

\newpage
\section{Supplementary Results}
\label{appendixA}

Here, we present the detailed information of the eight real-world datasets discussed in the main manuscript, as summarized in Table \ref{tab:tab3}.

\begin{table*}[ht!]
\centering
\caption{Properties and statistics of eight real-world datasets.}
\begin{tabular}{ccccccccc|c}
\toprule
\textbf{Dataset} & \textbf{Texas} & \textbf{Wisconsin} & \textbf{Cohameleon} & \textbf{Cornell} & \textbf{Citeseer} & \textbf{PubMed} & \textbf{Cora} & \textbf{Photo} & \textbf{OGB-ARXIV}\\
\midrule
Classes     & 5    & 5  & 5  & 5    & 7    & 3     & 6  & 8  & 49\\
Features    & 1,703 & 1,703 & 235 & 1,703 & 3,703 & 500   & 1,433 & 745 & 128\\
\#Nodes     & 183  & 183 & 2277 & 183  & 2,120 & 19,717 & 2,485 & 7487 & 169343\\
Edges       & 309  & 466 & 31371 & 277  & 3,679 & 44,324 & 5,069 & 119043 & 1166243 \\
Hom. ratio  & 0.11 & 0.21 & 0.23 & 0.30 & 0.74 & 0.79  & 0.81 & 0.83 & N/A\\
\bottomrule
\end{tabular}

\label{tab:tab3}
\end{table*}

\subsection{Datasets:}
\quad\textbf{Texas:} The Texas dataset is a small-scale heterophilic graph derived from university webpages. Nodes represent web pages, and edges denote hyperlinks between them. Each node is associated with a bag-of-words feature vector, and the task is to classify web pages into categories\footnote{\url{http://www.cs.cmu.edu/afs/cs.cmu.edu/project/theo-11/www/wwkb/}}.

\textbf{Wisconsin:} The Wisconsin dataset is similar to Texas, comprising university web pages and their hyperlinks, with bag-of-words node features. The classification task involves identifying the categories of web pages.

\textbf{Chameleon:} Chameleon is subgraph of web pages in Wikipedia. The node in Wikipedia graphs represent web pages, the edge mean mutual links between pages, and the node feature corresponds to several informative nouns in the Wikipedia page. All nodes are classified into 5 categories based on the average monthly traffic \cite{squirrel}.

\textbf{Cornell:} The Cornell dataset, another heterophilic graph, consists of web pages from Cornell University and their interconnections. Nodes are described by bag-of-words features, and the goal is to classify web pages. Its structural characteristics and small size make it a challenging benchmark for GNNs.

\textbf{Citeseer:} The Citeseer dataset is a citation network where nodes represent scientific publications and edges correspond to citation relationships. Each node is associated with a bag-of-words representation of the paper's abstract, and the task is to classify papers into predefined research topics \cite{citeseer}.

\textbf{Pubmed:} PubMed is a large citation network where nodes represent biomedical papers, and edges signify citation links. Node features are derived from Term Frequency-Inverse Document Frequency (TF-IDF) representations of abstracts. The classification task involves assigning each paper to one of three predefined categories \cite{pubmed}.

\textbf{Cora:} The Cora dataset is a widely-used citation network where nodes correspond to research papers and edges represent citation relationships. Each node is described by a bag-of-words feature vector derived from the paper's content. The task is to classify papers into one of seven machine learning-related topics \cite{cora}. 

\textbf{Photo:} Amazon Photo is a product recommendation graph derived from the Amazon product review data. In this dataset, nodes represent individual products, and edges signify co-purchases or similarity between products based on customer behavior. Each node is associated with product features such as textual descriptions, categories, and user ratings \cite{photo}. 

\textbf{OGB-Arxiv:} The ogbn-arxiv dataset is a directed graph, representing the citation network between all Computer Science (CS) ARXIV papers indexed by MAG. Each node is an ARXIV paper and each directed edge indicates that one paper cites another one. Each paper comes with a 128-dimensional feature vector obtained by averaging the embeddings of words in its title and abstract\cite{ogb}.

\subsection{Baselines:}
In addition to earlier results comparison with typical GNNs, in Table \ref{tab:tab4} we conducted a more comprehensive comparison with over thirty diverse baselines across multiple categories:

\textbf{Classic GNN and deep learning models:} MLP \cite{mlp}, GCN \cite{gcn}, GAT \cite{gat}, GraphSage \cite{graphsage}, ChebNet \cite{chebnet}, SGC \cite{sgc}, GPN \cite{gpn}, MM-FGCN \cite{fgcn} and DGMAE\cite{dgmae}.

\textbf{Models Designed for Heterophilic Graphs:} Geom-GCN \cite{geom}, PloyGCL \cite{ploy}, MixHop \cite{mixhop}, H2GCN \cite{h2gcn}, GGCN \cite{ggcn}, LINX \cite{linkx}, GloGNN \cite{Glognn}, ACM-GCN \cite{acm-gcn}, WRGAT \cite{wrgat}, GPR-GNN \cite{gprgnn}, FA-GCN \cite{fa-gcn}, BEC-GNN\cite{becgnn} and GRAIN\cite{grain}.

\textbf{Methods Addressing Oversmoothing:} GCNII \cite{gcnii}, JKNet \cite{jknet}, PairNorm \cite{pairnorm}, GCON \cite{gcon} and Adaptive IRC\cite{irc}.

\textbf{Continuous-Time GNN Models:} CGNN \cite{cgnn}, GDE \cite{gde} and FROND \cite{frond}. 

\textbf{Diffusion-Based GNN Methods:} GRAND \cite{grand}, BLEND \cite{blend}, Sheaf \cite{sheaf}, GRAFF \cite{graff}, CDE \cite{cde}, GREAD \cite{gread}, ACMP \cite{acmp} and SDMG \cite{sdmg}.

\begin{table*}[ht!]
\centering
\setlength{\tabcolsep}{1.7pt} 
\renewcommand{\arraystretch}{1.2} 
\caption{Results on eight real-world datasets: mean $\pm$ std. dev. accuracy for 10 different data splits. Different types of baselines are separated by dividing lines in the table. We show the best three results in \textcolor{red}{\textbf{bold}}(first), \textcolor{cyan}{blue}(second), \textcolor{violet}{purple}(third).}
\setlength{\tabcolsep}{6.4pt} 
\begin{tabular*}{0.92\textwidth}{ccccccccccc}
\toprule
\textbf{Dataset} & \textbf{Texas} & \textbf{Wisconsin} & \textbf{Chameleon} & \textbf{Cornell} & \textbf{Citeseer} & \textbf{PubMed} & \textbf{Cora} & \textbf{Photo} \\
Hom. ratio  & 0.11 & 0.21 & 0.23 & 0.30 & 0.74 & 0.79  & 0.81 & 0.83 \\
\midrule
MLP\cite{mlp}      & 80.81$\pm$\tiny 4.75 & 85.29$\pm$\tiny 3.31 & 46.21$\pm$\tiny 2.99 & 81.89$\pm$\tiny 6.40 & 74.02$\pm$\tiny 1.90 & 75.69$\pm$\tiny 2.00 & 87.16$\pm$\tiny 0.37 & 88.45$\pm$\tiny 0.57\\
GCN\cite{gcn}      & 55.14$\pm$\tiny 5.16 & 51.76$\pm$\tiny 3.06 & 64.82$\pm$\tiny 2.24 & 60.54$\pm$\tiny 5.30 & 76.50$\pm$\tiny 1.36 & 88.42$\pm$\tiny 0.50 & 86.98$\pm$\tiny 1.27 & 91.20$\pm$\tiny 1.20 \\
GAT\cite{gat}      & 52.16$\pm$\tiny 6.63 & 49.41$\pm$\tiny 4.09 & 60.26$\pm$\tiny 2.50 & 61.89$\pm$\tiny 5.05 & 76.55$\pm$\tiny 1.23 & 87.30$\pm$\tiny 1.10 & 86.33$\pm$\tiny 0.48 & 85.70$\pm$\tiny 1.14 \\
GraphSage\cite{graphsage} & 82.43$\pm$\tiny 6.14 & 81.18$\pm$\tiny 5.56 & 58.73$\pm$\tiny 1.68 & 75.95$\pm$\tiny 5.01 & 76.04$\pm$\tiny 1.30 & 88.45$\pm$\tiny 0.50 & 86.90$\pm$\tiny 1.04 & 91.40$\pm$\tiny 1.30 \\
ChebNet\cite{chebnet}  & 78.37$\pm$\tiny 6.04 & 79.02$\pm$\tiny 3.18 & 58.64$\pm$\tiny 1.64 & 75.68$\pm$\tiny 6.94 & 75.07$\pm$\tiny 1.25 & 89.00$\pm$\tiny 0.46 & 85.45$\pm$\tiny 1.58 & / \\
SGC\cite{sgc}      & 58.10$\pm$\tiny 4.20 & 55.29$\pm$\tiny 4.28 & 42.45$\pm$\tiny 3.82 & 60.00$\pm$\tiny 3.59 & 76.01$\pm$\tiny 1.31 & 86.90$\pm$\tiny 1.32 & 86.12$\pm$\tiny 1.44 & / \\
GPN\cite{gpn}      & 79.95$\pm$\tiny 5.62 & 81.29$\pm$\tiny 5.09 & 70.90$\pm$\tiny 1.51 & 77.93$\pm$\tiny 5.93 & 77.02$\pm$\tiny 1.56 & 89.61$\pm$\tiny 0.36 & 88.07$\pm$\tiny 0.87 & / \\
MM-FGCN\cite{fgcn}  & 86.10$\pm$\tiny 4.50 & \textcolor{violet}{88.50$\pm$\tiny 4.10} & \textcolor{red}{\textbf{73.97$\pm$\tiny 2.10}} & \textcolor{cyan}{88.90$\pm$\tiny 8.30} & 73.90$\pm$\tiny 0.60 & 80.70$\pm$\tiny 0.20 & 84.40$\pm$\tiny 0.50 & 93.55$\pm$\tiny 1.68 \\
DGMAE\cite{dgmae}  & 88.11\,$\pm$\,{\tiny 5.16} & 88.43$\pm${\tiny 3.56} & 75.50$\pm${\tiny 1.17} & 78.65$\pm${\tiny 4.59} & 73.82$\pm${\tiny 0.64} & 81.10$\pm${\tiny 0.42} & 84.93$\pm${\tiny 0.51} & \textcolor{violet}{93.96$\pm${\tiny 0.41}} \\\midrule

Geom-GCN\cite{geom} & 66.76$\pm$\tiny 2.72 & 64.51$\pm$\tiny 3.66 & 60.00$\pm$\tiny 2.81 & 60.54$\pm$\tiny 3.67 & \textcolor{cyan}{78.02$\pm$\tiny 1.15} & 89.95$\pm$\tiny 0.47 & 85.35$\pm$\tiny 1.57 & 92.35$\pm$\tiny 1.31\\
PloyGCL\cite{ploy}  & 88.03$\pm$\tiny 1.80 & 85.50$\pm$\tiny 1.88 &  \textcolor{violet}{71.62$\pm$\tiny 0.96} & 82.62$\pm$\tiny 3.11 & \textcolor{red}{\textbf{79.81$\pm$\tiny 0.85}} & 87.57$\pm$\tiny 0.62 & 87.15$\pm$\tiny 0.27 & / \\
MixHop\cite{mixhop}   & 77.84$\pm$\tiny 7.73 & 75.88$\pm$\tiny 4.90 & 60.50$\pm$\tiny 2.51 & 73.51$\pm$\tiny 6.34 & 76.26$\pm$\tiny 1.33 & 85.31$\pm$\tiny 0.61 & 87.61$\pm$\tiny 0.85 & 93.60$\pm$\tiny 1.37 \\
H2GCN\cite{h2gcn}    & 84.86$\pm$\tiny 7.23 & 87.65$\pm$\tiny 4.98 & 60.11$\pm$\tiny 2.15 & 82.70$\pm$\tiny 5.28 & 77.11$\pm$\tiny 1.57 & 89.49$\pm$\tiny 0.38 & 87.87$\pm$\tiny 1.20 & / \\
GGCN\cite{ggcn}     & 84.86$\pm$\tiny 4.55 & 86.83$\pm$\tiny 3.29 & 71.14$\pm$\tiny 1.84 & 85.68$\pm$\tiny 6.63 & \textcolor{violet}{77.41$\pm$\tiny 1.65} & 89.15$\pm$\tiny 0.37 & 87.95$\pm$\tiny 1.05 & / \\
LINX\cite{linkx}     & 74.60$\pm$\tiny 3.33 & 75.49$\pm$\tiny 5.72 & 68.42$\pm$\tiny 1.38 & 77.84$\pm$\tiny 5.81 & 73.19$\pm$\tiny 0.99 & 89.62$\pm$\tiny 0.35 & 88.31$\pm$\tiny 1.13 & / \\
GloGNN\cite{Glognn}   & 84.32$\pm$\tiny 4.15 & 87.06$\pm$\tiny 3.53 & 69.78$\pm$\tiny 2.42 & 83.51$\pm$\tiny 4.26 & 77.31$\pm$\tiny 1.48 & 89.62$\pm$\tiny 0.45 & 87.15$\pm$\tiny 1.43 & / \\
ACM-GCN\cite{acm-gcn}  & 87.84$\pm$\tiny 4.40 & 88.43$\pm$\tiny 3.22 & 66.93$\pm$\tiny 1.85 & 85.14$\pm$\tiny 6.07 & 77.32$\pm$\tiny 1.70 & 90.00$\pm$\tiny 0.52 & 87.91$\pm$\tiny 0.95 & / \\
WRGAT\cite{wrgat}    & 83.62$\pm$\tiny 5.50 & 86.98$\pm$\tiny 3.78 & 65.24$\pm$\tiny 0.87 & 81.62$\pm$\tiny 3.90 & 76.81$\pm$\tiny 1.89 & 89.29$\pm$\tiny 0.38 & 88.20$\pm$\tiny 2.26 & / \\
GPR-GNN\cite{gprgnn}  & 78.38$\pm$\tiny 4.36 & 82.94$\pm$\tiny 4.21 & 46.58$\pm$\tiny 1.84 & 80.27$\pm$\tiny 8.11 & 77.13$\pm$\tiny 1.67 & 87.54$\pm$\tiny 0.38 & 87.95$\pm$\tiny 1.18 & / \\
FA-GCN\cite{fa-gcn}   & 82.43$\pm$\tiny 6.89 & 82.94$\pm$\tiny 7.95 & 55.22$\pm$\tiny 1.39 & 79.19$\pm$\tiny 9.79 & 76.87$\pm$\tiny 1.56 & 87.45$\pm$\tiny 0.61 & 87.21$\pm$\tiny 1.43 & / \\
    BEC-GNN\cite{becgnn} & 88.23$\pm${\tiny 1.75} & 89.55$\pm${\tiny 1.22} & /                          & \textcolor{red}{\textbf{93.82$\pm${\tiny 1.10}}} & \textcolor{red}{\textbf{80.43$\pm${\tiny 0.88}}} & 89.40$\pm${\tiny 0.19} & 88.50$\pm${\tiny 1.32} & /     \\
GRAIN\cite{grain}   & 87.69$\pm${\tiny 0.00} & 89.01$\pm${\tiny 0.00} & 56.43$\pm${\tiny 0.00} & \textcolor{red}{\textbf{90.12$\pm${\tiny 0.00}}} & \textcolor{red}{\textbf{81.25$\pm${\tiny 0.00}}} & 87.04$\pm${\tiny 0.00} & 88.52$\pm${\tiny 0.00} & /     \\
\midrule
GCNII\cite{gcnii}    & 77.57$\pm$\tiny 3.83 & 80.39$\pm$\tiny 3.40 & 63.86$\pm$\tiny 3.04 & 77.86$\pm$\tiny 3.79 & 77.33$\pm$\tiny 1.48 & \textcolor{violet}{90.15$\pm$\tiny 0.43} & \textcolor{violet}{88.37$\pm$\tiny 1.25} & / \\
JKNet\cite{jknet}    & 62.70$\pm$\tiny 8.34 & 53.14$\pm$\tiny 5.22 & 52.63$\pm$\tiny 3.90 & 59.72$\pm$\tiny 4.60 & 75.99$\pm$\tiny 1.28 & 87.23$\pm$\tiny 0.55 & 86.48$\pm$\tiny 1.04 & / \\
PairNorm\cite{pairnorm} & 60.27$\pm$\tiny 4.34 & 48.43$\pm$\tiny 6.14 & 62.74$\pm$\tiny 2.82 & 58.92$\pm$\tiny 3.15 & 73.59$\pm$\tiny 1.47 & 87.53$\pm$\tiny 0.44 & 85.79$\pm$\tiny 1.01 & / \\
GCON\cite{gcon}     & 85.40$\pm$\tiny 4.20 & 87.80$\pm$\tiny 3.30 & 48.31$\pm$\tiny 1.53 & 84.30$\pm$\tiny 4.80 & 76.46$\pm$\tiny 1.70 & 87.71$\pm$\tiny 0.35 & 87.40$\pm$\tiny 1.82 & / \\
Adaptive IRC\cite{irc}
& 84.60$\pm${\tiny 6.1}
& 82.20$\pm${\tiny 2.3}
& 65.00$\pm${\tiny 2.0}
& 72.40$\pm${\tiny 5.7}
& 70.20$\pm${\tiny 0.4}
& 77.40$\pm${\tiny 0.6}
& 80.70$\pm${\tiny 0.4}
& / \\
\midrule
CGNN\cite{cgnn}     & 71.35$\pm$\tiny 4.05 & 74.31$\pm$\tiny 7.26 & 46.89$\pm$\tiny 1.66 & 66.22$\pm$\tiny 7.69 & 76.91$\pm$\tiny 1.18 & 87.70$\pm$\tiny 0.49 & 87.10$\pm$\tiny 1.35 & / \\
GDE\cite{gde}      & 74.05$\pm$\tiny 6.96 & 79.80$\pm$\tiny 5.62 & 87.22$\pm$\tiny 1.41 & 82.43$\pm$\tiny 7.07 & 76.21$\pm$\tiny 2.11 & 87.80$\pm$\tiny 0.38 & 87.22$\pm$\tiny 1.41 & 91.30$\pm$\tiny 1.50 \\
FROND\cite{frond}    & 75.56$\pm$\tiny 5.15 & 77.95$\pm$\tiny 6.75 & 71.45$\pm$\tiny 1.98 & 75.36$\pm$\tiny 6.19 & 74.70$\pm$\tiny 1.90 & 79.40$\pm$\tiny 1.50 & 84.80$\pm$\tiny 1.10 & 93.10$\pm$\tiny 1.50 \\
\midrule
GRAND\cite{grand}    & 75.68$\pm$\tiny 1.25 & 79.41$\pm$\tiny 4.12 & 54.67$\pm$\tiny 2.54 & 82.46$\pm$\tiny 7.09 & 76.46$\pm$\tiny 1.77 & 89.02$\pm$\tiny 0.35 & 87.36$\pm$\tiny 0.96 & 92.30$\pm$\tiny 0.90 \\
BLEND\cite{blend}    & 83.24$\pm$\tiny 4.55 & 84.12$\pm$\tiny 3.56 & 60.11$\pm$\tiny 2.09 & 85.95$\pm$\tiny 6.82 & 76.63$\pm$\tiny 1.30 & 89.24$\pm$\tiny 0.42 & 88.09$\pm$\tiny 1.22 & 93.50$\pm$\tiny 0.30 \\
Sheaf\cite{sheaf}    & 85.05$\pm$\tiny 5.51 & \textcolor{cyan}{89.41$\pm$\tiny 4.74} & 68.04$\pm$\tiny 1.58 & 84.86$\pm$\tiny 4.71 & 76.70$\pm$\tiny 1.57 & 89.49$\pm$\tiny 0.40 & 86.90$\pm$\tiny 1.13 & / \\
GRAFF\cite{graff}    & \textcolor{violet}{88.38$\pm$\tiny 3.53} & 87.45$\pm$\tiny 2.94 & 71.08$\pm$\tiny 1.75 & 83.24$\pm$\tiny 6.49 & 76.92$\pm$\tiny 1.70 & 88.95$\pm$\tiny 0.52 & 87.61$\pm$\tiny 0.96 & 92.59$\pm$\tiny 0.96 \\
CDE\cite{cde}      & 87.57$\pm$\tiny 3.24 & 87.84$\pm$\tiny 4.87 & 68.45$\pm$\tiny 2.47 & 86.22$\pm$\tiny 5.05 & 76.40$\pm$\tiny 1.37 & 86.68$\pm$\tiny 1.90 & 87.70$\pm$\tiny 1.66 & 93.05$\pm$\tiny 1.61 \\
GREAD\cite{gread}    & \textcolor{cyan}{88.92$\pm$\tiny 3.72} & \textcolor{cyan}{89.41$\pm$\tiny 1.30} & 71.38$\pm$\tiny 1.53 & \textcolor{violet}{87.03$\pm$\tiny 4.95} & \textcolor{violet}{77.60$\pm$\tiny 1.81} & \textcolor{cyan}{90.23$\pm$\tiny 0.55} & \textcolor{cyan}{88.57$\pm$\tiny 0.66} & 92.65$\pm$\tiny 1.11 \\
ACMP\cite{acmp}     & 86.20$\pm$\tiny 3.00 & 86.10$\pm$\tiny 4.00 & 52.63$\pm$\tiny 2.28 & 85.40$\pm$\tiny 7.00 & 75.50$\pm$\tiny 1.00 & 79.40$\pm$\tiny 0.40 & 84.90$\pm$\tiny 0.60 & 91.80$\pm$\tiny 1.10 \\
SDMG\cite{sdmg} & /& /& /& /& 73.90$\pm$  \tiny0.40 &	80.00$\pm$\tiny0.50	& 84.30$\pm$\tiny0.50 & \textcolor{cyan}{94.70$\pm$\tiny0.20}\\
\midrule

\textbf{GNSN}     & \textcolor{red}{\textbf{91.89$\pm$\tiny 2.95}} & \textcolor{red}{\textbf{90.02$\pm$\tiny 1.94}} &  \textcolor{cyan}{71.85$\pm$\tiny 1.56} & \textcolor{cyan}{89.19$\pm$\tiny 5.13}& 76.87$\pm$\tiny 1.23 & \textcolor{red}{\textbf{90.41$\pm$\tiny 0.58}} & \textcolor{red}{\textbf{88.73$\pm$\tiny 1.06}} & \textbf{\textcolor{red}{95.43$\pm$\tiny 1.26}} \\

\bottomrule
\end{tabular*}
\label{tab:tab4}
\end{table*}

\clearpage
\section{Supplementary Datasets}
\label{appendixB}
In addition to the eight real-world datasets discussed in the main manuscript, we introduce an additional set of seven datasets: two homophilic datasets (Coauthor CS and Computer) and two heterophilic datasets (Film and Squirrel). Detailed information about the supplementary datasets is provided in Table \ref{tab:tab5}. Furthermore, we compare the performance of our model against several state-of-the-art baselines on these datasets, as shown in Table \ref{tab:tab6}, Table \ref{tab:dataset_comparison} and Table \ref{tab:ogb_arxiv_results}.

\begin{table*}[h!t]
\centering
\caption{Dataset statistics for additional five datasets}
\begin{tabular}{cccccc}
\toprule
\textbf{Dataset} & \textbf{Film} & \textbf{Squirrel} & \textbf{Computer} & \textbf{Coauthor CS}\\
\midrule
Classes     & 5     & 5     & 10    & 15   \\
Features    & 932   & 2,089 & 767   & 6,805   \\
\#Nodes     & 7,600 & 5,201 & 13,381 & 18,333 \\
Edges       & 26,752 & 198,353 & 245,778 & 81,894 \\
Hom. ratio  & 0.22  & 0.22  & 0.77  & 0.80  \\
\bottomrule
\end{tabular}
\label{tab:tab5}
\end{table*}

\subsection{Homophilic datasets}

\quad\textbf{Computer:} Amazon Computers is a co-purchase graph extracted from Amazon, where nodes represent products, edges represent the co-purchased relations of products, and features are bag-of-words vectors extracted from product reviews \cite{photo}.

\textbf{Coauthor CS:} Coauthor CS is co-authorship graph based on the Microsoft Academic Graph from the KDD Cup 2016 challenge. Here, nodes are authors, that are connected by an edge if they co-authored a paper; node features represent paper keywords for each author’s papers, and class labels indicate most active fields of study for each author \cite{co}.

\textbf{Baselines:} On these additional two homophilic datasets, we performed a comprehensive comparison with the following baseline models: GCN \cite{gcn}, GAT and GAT-ppr3 \cite{gat}, MoNet \cite{monet}, GraphSage \cite{graphsage}, CGNN \cite{cgnn}, GDE \cite{gde}, FROND \cite{frond}, GRAND \cite{graff}, GRAND++ \cite{grand++}, BLEND \cite{blend} and ACMP \cite{acmp}.

\begin{table}[ht!]
\centering
\caption{Results on two supplementary homophilic datasets.}
\begin{tabular}{ccc}
\toprule
\textbf{Dataset} & \textbf{Computer} & \textbf{Coauthor CS} \\ 
\midrule
GCN\cite{gcn} &  82.60$\pm$2.40 & 91.10$\pm$0.50 \\ 
GAT\cite{gat} &  78.00$\pm$0.95 & 90.50$\pm$0.60 \\ 
GAT-ppr\cite{gat} & 85.40$\pm$0.10 & 91.30$\pm$0.10 \\ 
MoNet\cite{monet}  & 83.50$\pm$2.20 & 90.80$\pm$0.60 \\ 
GraphSage\cite{graphsage} & 82.40$\pm$1.80 & 91.30$\pm$4.80 \\ 
CGNN\cite{cgnn} & 80.20$\pm$4.20 & 92.30$\pm$0.20 \\ 
GDE\cite{gde}  & 82.90$\pm$0.60 & 91.60$\pm$0.10 \\ 
FROND\cite{frond} & 84.40$\pm$1.50 & \textcolor{cyan}{93.00$\pm$0.90} \\
\midrule
GRAND\cite{grand}  & 83.70$\pm$1.20 & \textcolor{violet}{92.90$\pm$0.40} \\ 
GRAND++\cite{grand++} & \textcolor{violet}{85.73$\pm$0.50} & 90.80$\pm$0.34 \\
BLEND\cite{blend} & \textcolor{cyan}{86.90$\pm$0.60} & \textcolor{violet}{92.90$\pm$0.20} \\
ACMP\cite{acmp} & 84.40$\pm$1.60 & \textcolor{cyan}{93.00$\pm$0.50} \\  
\midrule
\textbf{GNSN} & \textcolor{red}{\textbf{88.13$\pm$1.22}} \scriptsize(1.23\(\uparrow\)) & \textcolor{red}{\textbf{93.31$\pm$0.65}} \scriptsize(0.31\(\uparrow\)) \\
\bottomrule
\label{tab:tab6}
\end{tabular}%
\end{table}

\newpage
\subsection{Heterophilic datasets}

\quad\textbf{Film:} The Film dataset is a subgraph of the film-director-actor-writer network. Each node represents an actor, and edges between nodes denote co-occurrence relationships found on Wikipedia pages. The node features correspond to keywords extracted from these pages. The dataset classifies the nodes into five categories based on the type of actors \cite{film}.

\textbf{Squirrel:} Squirrel is collected from the English Wikipedia. These datasets represent page-page networks on specific topics (chameleons, crocodiles and squirrels). Nodes represent articles and edges are mutual links between them. The features json files contain the features of articles - each key is a page id, and node features are given as lists. The presence of a feature in the feature list means that an informative noun appeared in the text of the Wikipedia article \cite{squirrel}.

\textbf{Baselines:} On these additional two heterophilic datasets, we performed a comprehensive comparison with the following baseline models: Geom-GCN \cite{geom}, H2GCN \cite{h2gcn}, GGCN \cite{ggcn}, LINX \cite{linkx}, GloGNN \cite{Glognn}, ACM-GCN \cite{acm-gcn}, GRAND \cite{grand}, BLEND \cite{blend}, Sheaf \cite{sheaf}, GRAFF \cite{graff} and GREAD \cite{gread}.

\begin{table}[h]
\centering
\caption{Results on two supplementary heterophilic datasets.}
\label{tab:dataset_comparison}
\begin{tabular}{ccc}
\toprule
\textbf{Dataset} & \textbf{Film} & \textbf{Squirrel} \\
\midrule
Geom-GCN\cite{geom}   & 31.59$\pm$1.15 & 38.15$\pm$0.92 \\
H2GCN\cite{h2gcn}      & 35.70$\pm$1.00 & 36.48$\pm$1.86 \\
GGCN\cite{ggcn}       & 37.54$\pm$1.56 & 55.17$\pm$1.58 \\
LINX\cite{linkx}       & 36.10$\pm$1.55 & 61.81$\pm$1.80 \\
GloGNN\cite{Glognn}     & 37.35$\pm$1.30 & \textcolor{violet}{57.54$\pm$1.39} \\
ACM-GCN\cite{acm-gcn}    & 36.28$\pm$1.09 & 54.40$\pm$1.88 \\ 
GRAND\cite{grand}      & 35.62$\pm$1.01 & 40.05$\pm$1.50 \\
BLEND\cite{blend}      & 35.63$\pm$1.01 & 43.06$\pm$1.39 \\
Sheaf\cite{sheaf}      & \textcolor{violet}{37.81$\pm$1.15} & 56.34$\pm$1.32 \\
GRAFF\cite{graff}      & 36.09$\pm$0.81 & 54.52$\pm$1.37 \\ 
GREAD\cite{gread}      & \textcolor{cyan}{37.90$\pm$1.17} & \textcolor{cyan}{59.22$\pm$1.44} \\ \hline
\textbf{GNSN}       & \textcolor{red}{\textbf{38.88$\pm$0.82}} \scriptsize(0.98\(\uparrow\)) & \textcolor{red}{\textbf{59.54$\pm$1.67}} \scriptsize(0.34\(\uparrow\)) \\
\bottomrule
\end{tabular}
\end{table}

\newpage
\section{Detailed Experimental Configuration}
\label{appendixC}
\subsection{Experimental environments}
The experiments were conducted in the following software and hardware environments: 

\begin{table}[h]
\centering
\caption{Experimental environments}
\begin{tabular}{cc}
\toprule
\textbf{Components}         & \textbf{Version}      \\
\midrule
CPU               & 13th i7      \\
GPU               & RTX 3080     \\
GPU Memory        & 10GB         \\
CUDA              & 11.3         \\
Driver            & 550.12       \\
Operating system  & Ubuntu 22.04 \\
Python            & 3.8.13       \\
Pytorch           & 1.10.2       \\
Pytorch Geometric & 2.0.4        \\
Torchdiffeq       & 0.2.2        \\
Numpy             & 1.21.1       \\
Scipy             & 1.10.1       \\
Scikit-learn      & 1.2.1        \\
WandB             & 0.15.5       \\
Matplotlib        & 3.5.1       \\
\bottomrule
\end{tabular}
\label{tab:tab8}
\end{table}

\newpage

\section{Inspiration \& Derivation}
\label{appendixD}

\subsection{From scalar flux towards message flow}
\quad\textbf{Scalar flux in fluid system}

In a dynamic fluid system, for a scalar field \(c(x,t)\) (such as concentration or heat), the total flux \(J\) consists of both convection and diffusion terms.

The diffusion flux is determined by the gradient of the scalar field:
\begin{align}
    \overrightarrow{J}_\mathrm{diff} = -D \nabla c, \label{eq:eq16}
\end{align}
where \(D\) is the diffusion coefficient and \(\nabla c\) is the gradient of the scalar field.

The convection flux is determined by the velocity field \(\mathbf{u}\) and the scalar field:
\begin{align}
    \overrightarrow{J}_\mathrm{conv} = \mathbf{u}c, \label{eq:eq17}
\end{align}
where \(\mathbf{u}\) is the velocity field, which represents the direction and rate of scalar flow, and \(c\) is the value of the scalar field.

Thus, the total flux \(\overrightarrow{J}\) is the combination of both diffusion and convection:
\begin{align}
     \overrightarrow{J} = \overrightarrow{J}_\mathrm{diff} + \overrightarrow{J}_\mathrm{conv} = -D \nabla c + \mathbf{u}c, \label{eq:eq18}
\end{align}
Using the control-volume method, the inflow and outflow of the scalar quantity can be determined by performing an integration over the surface of a control volume. Let \(V\) denote a small control volume, and \(\partial V \) represent its boundary surface. The flux through the surface of the control volume can be formulated as the surface integral of the flux vector \(\overrightarrow{J}\), expressed as:
\begin{align}
    \mathrm{Outflux-Influx} = \int_{\partial V }^{} \overrightarrow{J} \cdot \mathbf{n} \mathrm{d}A, \label{eq:eq19}  
\end{align}
where \(\mathbf{n}\) is the unit normal vector to the surface element \(\mathrm{d}A\) (pointing outward from the control volume), \(\mathrm{d}A\) is the infinitesimal surface area element of \(\partial V\).

Substituting the total scalar flux \(\overrightarrow{J}\), we get:
\begin{align}
    \mathrm{Outflux-Influx} = \int_{\partial V }^{} (-D \nabla c + \mathbf{u}c) \cdot \mathbf{n} \mathrm{d}A. \label{eq:eq20}  
\end{align}
The first term, \(\int_{\partial V }^{} -D \nabla c \cdot \mathbf{n} \mathrm{d}A\), represents the contribution from the diffusion flux, describing the diffusion flow of the scalar field. The second term, \(\int_{\partial V }^{} \mathbf{u}c \cdot \mathbf{n} \mathrm{d}A\), represents the contribution from the convection flux, describing the transport of the scalar field through the flow. This integral represents the total flux passing through the boundary of the control volume, which accounts for both the inflow and outflow of the scalar quantity. By appropriately substituting the flux \(\overrightarrow{J}\) in terms of the diffusion and convection contributions, the overall flux through the control volume can be evaluated.

\textbf{Message flow on graph}

In a graph \(\mathcal{G}={\mathcal{(V,E)}}={\mathbf{(A,X)}}\), the message flow at a node \(i\) is determined by the aggregated inflow and outflow of message through its edges. Let \(\mathbf{x}(i)\) denote the feature vector of node \(i\), and let \(\mathbf{x}(j)\) represent the feature vector of a neighboring node \(j\). The message flow of node \(i\) can be divided into the following two components:

\textbf{\(\bullet\) Inflow:} Messages transmitted from neighboring nodes \(j\) to node \(i\) via edges. This message flow typically depends on the node feature \(\mathbf{x}(i)\) and  \(\mathbf{x}(j)\), the edge features \(e_{i,j}\), and the edge weight or adjacency relation \(a_{i,j}\).

\textbf{\(\bullet\) Outflow:} Messages flowed out from node \(i\) to its neighboring nodes.

Since message passing on graph structures does not follow a conservation law analogous to mass conservation in physical systems, we define the message flow for node \(i\) solely based on inflow during the message-passing process \cite{flowx, f2gnn}. Given this consideration, the message flow for node \(i\) at the current time step is defined as the sum of inflows from all its neighbors \(j \in \mathcal{N}_{i}\), expressed as: 
\begin{align}
    \mathrm{Message\;flow}(i) =\sum_{j \in \mathcal{N}_{i}}e_{i,j}\mathbf{m}_{i,j}, \label{eq:eq21}
\end{align}
where \(\mathbf{m}_{i,j}\) is the message transmitted from node \(j\) to node \(i\). Combining Eq.\ref{eq:eq4}, which describes the message passing and aggregation-update process for node \(i\), the expression for \(\mathbf{m}_{i,j}\) can be derived as follows:
\begin{align}
    \mathbf{m}_{i,j} = \zeta(\mathbf{x}_i, \mathbf{x}_{j}, \mathbf{A}_{i,j}), \label{eq:eq22}
\end{align}
where \(\zeta\) is a differentiable function.

\textbf{Finite-Volume Analogy}

In continuous fluid dynamics, scalar flux represents the transport of quantities across spatial boundaries, governed by the continuum assumption. In the graph domain, however, the structure is inherently discrete, lacking continuous spatial coordinates. To bridge this gap, we adopt a \textit{Finite Volume Method (FVM)} perspective to model message passing. Under this view, the graph is treated as a discretized system where:

\begin{itemize}
\item \textbf{Nodes-Control Volumes:} Each node $i$ functions as a discrete control volume, responsible for accumulating and updating scalar features (mass/energy).
\item \textbf{Edges-Fixed Interfaces:} The edges connecting nodes act as the boundary interfaces. Crucially, unlike continuous fluids where flow direction is arbitrary, the transport direction on graphs is \textit{strictly constrained by the topology} (i.e., along the edges).
\item \textbf{Message Flow-Scalar Flux:} The aggregation of messages from neighbors is analogous to the net scalar flux entering the control volume through its interfaces.
\end{itemize}

By analogy with the total scalar flux in a transport equation, we hypothesize that the message flow for node $i$ can be decomposed into two fundamental components:
\begin{align}
    \mathrm{Message\;flow}(i) = J_{\mathrm{conv},i} + J_{\mathrm{diff},i}, \label{eq:eq23}
\end{align}
Here:

\begin{itemize}
    \item $J_{\mathrm{diff},i}$ represents the \textbf{diffusive component}, driven by the feature gradient between connected nodes, promoting local smoothing and homogenization.
    \item $J_{\mathrm{conv},i}$ represents the \textbf{convective component}, describing the directed transport of features. Since the direction is fixed by the edge, this component is governed by a node-specific convective intensity (or effective velocity), which determines the magnitude of information throughput.
\end{itemize}

\subsection{Graph diffusion \& convection}
\quad\textbf{Diffusion on graph}

Referring to works such as GRAND \cite{grand}, GRAND++ \cite{grand++}, GREAD \cite{gread} and ACMP \cite{acmp}, it is evident that the diffusion equation effectively models message passing on graphs. The diffusion flow represents the process of local smoothing driven by the feature gradients between nodes. This mechanism captures how message propagates through the graph, balancing feature differences across connected nodes and leading to the homogenization of features within local neighborhoods.

Building upon prior diffusion-based works and Eq.2, the diffusion flow of node \(i\) can be formulated as:
\begin{align}
    J_{\mathrm{diff},i} = \sum_{j \in \mathcal{N}_{i}}D \mathbf{A}_{i,j}(\mathbf{x}_{i}-\mathbf{x}_j),\label{eq:eq24}
\end{align}
where \(D\) is a diffusion coefficient modulating the propagation strength, with its exact form yet to be explicitly defined, and \( \mathbf{A}_{i,j}\) still is the adjacency relation.

\textbf{Convection on graph}

To theoretically ground the finite-volume perspective introduced in Section 3.2, we formalize the convection mechanism on graphs. In continuous fluid dynamics, convective flux represents the directed transport of mass driven by a velocity field. On graphs, however, we lack continuous spatial coordinates. Consistent with the main text, we adopt a topology-constrained view where edges act as fixed interaction channels.

We define the convective message flow for node $i$ by adapting the classical convection term (Eq.\ref{eq:eq17}) to the graph domain:
\begin{align}
    J_{\mathrm{conv},i} = \mathbf{u}_i \sum_{j \in \mathcal{N}_{i}} \mathbf{A}_{i,j} \cdot \mathbf{h}_{j}, \label{eq:eq25}
\end{align}
Here, $\mathbf{u}_i \in \mathbb{R}^+$ corresponds to the effective convection velocity defined in Proposition \ref{pro:velo}. 

It is critical to clarify the physical nature of $\mathbf{u}_i$ in this equation:
\begin{itemize}
    \item \textbf{Scalar Intensity vs. Spatial Vector:} Unlike physical velocity vectors that dictate flow direction in Euclidean space, the directionality in Eq. (\ref{eq:eq25}) is inherently determined by the adjacency structure ($\mathbf{A}_{i,j}$). Thus, $\mathbf{u}_i$ does not represent a kinematic vector. Instead, as detailed in Section 3.2, it is a scalar quantity representing the transport intensity.
    \item \textbf{Dynamic Permeability:} Physically, $\mathbf{u}_i$ acts like a learnable, dynamic permeability coefficient for the control unit at node $i$. A high value of $\mathbf{u}_i$ indicates that the node is in a "high-flux" state, actively aggregating information from its interaction channels, effectively simulating the rapid transport characteristic of convection.
\end{itemize}

\textbf{Diffusion coefficient}

In Eq. \ref{eq:eq16}, \(D\) is a diffusion coefficient modulating the propagation strength. When mapped to a graph, \(D\) represents the similarity between nodes. \(\mathbf{A}_{i,j}\)is the \((i,j)\)-th element of the adjacency matrix \(\mathbf{A}\), representing the adjacency relationship between nodes \(i\) and \(j\). Under general conditions, the diffusion coefficient \(D\) is not constant but exhibits anisotropic and time-varying properties. On the graph, \(D\) can be associated with a node-related vector, with shape \(n\times n\) , where \(n\) is the number of nodes. Considering that \(D\) and \(\mathbf{A}\) share the same shapes, they can be integrated into \(\widetilde{\mathbf{A}}\), a comprehensive diffusion coefficient, where \cite{cde,grand,gread}:
\begin{align*}
\widetilde{\mathbf{A} }=f_{\bowtie }(D,\mathbf{A})
\end{align*}
\begin{align}
  {\mathbf{\widetilde{A}}}_{i, j}:=\operatorname{softmax}\left(\frac{\left(\mathbf{W}_{Q} \mathbf{h}_{i}\right)^{\top} \mathbf{W}_{K} \mathbf{h}_{j}}{d_{K}}\right).
\end{align}
thus, 
\(\mathbf{\widetilde{A}}\), while representing the adjacency relation between nodes, also encodes the similarity between them. Consequently, the diffusion flow \(J_{\mathrm{diff},i}\) in Eq.\ref{eq:eq25} can be simplified as follows:
\begin{align}
    J_{\mathrm{diff},i} = \sum_{j \in \mathcal{N}_{i}}\mathbf{\widetilde{A}}_{i, j}(\mathbf{x}_{i}-\mathbf{x}_j).\label{eq:eq27}
\end{align}

\textbf{Velocity Field Construction via Inverse Inference}

In classical fluid mechanics, the velocity field $\mathbf{u}$ is a fundamental physical quantity that actively drives convection flux. A particle's displacement is a consequence of this velocity. However, in the graph domain, no such pre-defined physical velocity exists. To overcome this, we adopt an inverse inference approach: instead of prescribing a velocity field in a physical field, we infer the effective convective intensity from the magnitude of the observed feature flux.

\textbf{The Time-Averaged Perspective:} Consider a fluid particle moving over a time interval $T$. Its average speed is defined as the total distance traveled divided by $T$. Analogously, in our continuous-time GNN framework (modeled via Neural ODEs), the "depth" of the network corresponds to the integration time $t$. The feature evolution follows:
\begin{align}
    \mathbf{h}(T) = \mathbf{h}(0) + \int_{0}^{T} \frac{\partial \mathbf{h}(t)}{\partial t} \mathrm{d}t
\end{align}
Within this continuous framework, we define the velocity $\mathcal{u}_i$ not as an instantaneous state, but as the time-averaged transport intensity over the evolution period.

\textbf{Segmented Time Mechanism:} We introduce a node-specific adaptive temporal window $[0, t_i] \subseteq [0, T]$ to capture the local dynamics of convection. The evolution is conceptually divided into:
\begin{itemize}
    \item \textbf{Active Phase $[0, t_i]$:} The convection mechanism is active, contributing to the "effective displacement" (feature transport) of the node.
    \item \textbf{Dormant Phase $(t_i, T]$:} The convection contribution saturates or becomes negligible.
\end{itemize}

\textbf{Mathematical Formulation:} We define the effective velocity $\mathbf{u}_i$ as the time-averaged magnitude of the convective flux accumulated during the active phase. Let $F_i(\tau) = \sum_{j \in \mathcal{N}(i)} \alpha_{ij}(\tau) \mathbf{h}_{j}(\tau)$ represent the instantaneous net feature flux entering node $i$. The velocity is formulated as:
\begin{align}
    \mathbf{u}_{i} = \frac{1}{T} \int_{0}^{T} \|F_{i}(\tau)\|_2 \cdot \mathbb{I}(\tau \leq t_i) \, d \tau = \frac{1}{T} \int_{0}^{t_{i}} \|F_{i}(\tau)\|_2 \, d \tau \label{eq:ui_def}
\end{align}
Here, $\mathbb{I}(\cdot)$ is the indicator function. The term $\Psi_i = \int_{0}^{t_{i}} \|F_{i}(\tau)\|_2 d \tau$ represents the cumulative convective impact. Dividing by the total system time $T$ normalizes this impact into a rate, i.e., an effective velocity.

\textbf{Properties and Structural Awareness:} This definition satisfies several key theoretical properties:
\begin{itemize}
    \item \textbf{Non-negativity:} Since the integrand is a vector norm $\| \cdot \|_2 \geq 0$, and $T > 0$, it strictly follows that $\mathbf{u}_i \in \mathbb{R}_{\geq 0}$. This aligns with the definition of $u_i$ as a scalar intensity magnitude.
    \item \textbf{Structure Awareness:} The term $F_i(\tau)$ explicitly aggregates neighbors based on the graph topology (via $\alpha_{ij}$). Thus, $u_i$ is not a free parameter but is structurally grounded, reflecting the local connectivity and feature distribution of the graph.
    \item \textbf{Monotonicity w.r.t. Active Horizon:} Let $\mathcal{I}(t) = \int_{0}^{t} \|F_{i}(\tau)\|_2 d \tau$. Since the integrand $\|F_{i}(\tau)\|_2$ is non-negative, the accumulation function $\mathcal{I}(t)$ is monotonically non-decreasing. Consequently, for any $t_i' > t_i$, we have $\mathbf{u}_i(t_i') \geq \mathbf{u}_i(t_i)$.
\end{itemize}

\textbf{Physical Interpretation:} The parameter $t_i$ serves as a convection horizon. A larger $t_i$ implies that the node sustains active information exchange for a longer duration, resulting in a higher effective velocity $\mathbf{u}_i$. Conversely, a small $t_i$ indicates that the node's convective transport is transient or weak. By learning or adapting $t_i$ (e.g., based on local feature convergence where $\|\frac{\partial \mathbf{h}}{\partial t}\| \to 0$), the model dynamically assigns high convection velocities to nodes that require long-range or intensive message passing.

\subsection{Total Message Flow Integration}

Having established the distinct mechanisms for diffusion (via the coefficient $\mathbf{\widetilde{A}}_{i,j}$) and convection (via the effective velocity $\mathbf{u}_i$), we can now formulate the unified transport equation on the graph. By combining the diffusive flux (Eq. \ref{eq:eq24}) and the convective flux (Eq. \ref{eq:eq25}), the total message flow entering node $i$ is expressed as:

\begin{align}
    \mathrm{Message\;flow}(i) &= J_{\mathrm{conv},i} + J_{\mathrm{diff},i} \nonumber = \underbrace{\mathbf{u}_{i} \sum_{j \in \mathcal{N}_{i}} \mathbf{A}_{i,j} \mathbf{h}_j}_{\text{Convection: Directed Transport}} + \underbrace{\sum_{j \in \mathcal{N}_{i}} \mathbf{\widetilde{A}}_{i,j} (\mathbf{h}_j - \mathbf{h}_i)}_{\text{Diffusion: Gradient Smoothing}}. \label{eq:eq30}
\end{align}
Note that we utilize the feature difference $(\mathbf{h}_j - \mathbf{h}_i)$ to represent the diffusive gradient driving flow from neighbor $j$ into node $i$.

\textbf{Functional Interpretation of Velocity.} 
While $\mathbf{u}_i$ is theoretically constructed via the time-integral of flux (as derived in the previous section), in the computational implementation of the Neural ODE, it manifests as a state-dependent modulation parameter. Since $\mathbf{u}_i$ is governed by the adaptive horizon $t_{i}$, which in turn is learned from the local node features and topology, we can formalize $\mathbf{u}_i$ as a functional of the graph state:
\begin{align}
    \mathbf{u}_i = \Phi(\mathbf{h}_i, \mathcal{N}_i; \theta),
\end{align}
where $\Phi$ represents the learnable mapping parameterized by $\theta$. Consequently, the total message flow in Eq. \ref{eq:eq30} can be generalized into the canonical Message Passing Neural Network (MPNN) framework:
\begin{align}
   \mathrm{Message\;flow}(i) = \sum_{j \in \mathcal{N}_{i}} \eta \left( \mathbf{h}_i, \mathbf{h}_{j}, \mathbf{A}_{i,j}; \mathbf{u}_i \right), \label{eq:eq31}
\end{align}
where $\eta$ is a differentiable message function. 

\textbf{Conclusion of Derivation:} Eq. \ref{eq:eq31} demonstrates that our GNSN architecture is not merely a heuristic improvement but a physically grounded instantiation of the message-passing paradigm. It explicitly decomposes the aggregation function $\eta$ into a \textbf{convective component} (governed by transport intensity $\mathbf{u}_i$) and a \textbf{diffusive component} (governed by coupling $\mathbf{\widetilde{A}}_{i,j}$). This formulation proves that the analogy from scalar flux in fluid dynamics to message flow on graphs is mathematically consistent, bridging fundamental physical intuition with rigorous GNN design.

\subsection{From Navier-Stokes Equation towards GNN}

The Navier-Stokes equations, deriving from conservation laws, serve as the cornerstone of fluid dynamics. To adapt this continuous physical framework to graph representation learning, we must identify the mathematical structure that best describes the evolution of node features.

\textbf{The Physics: Momentum vs. Transport:} The full Navier-Stokes framework consists of two coupled components:
\begin{itemize}
    \item \textbf{Momentum Equation (Navier-Stokes):} Describes the evolution of the velocity field $\mathbf{u}$ itself:
    \begin{align}
        \underbrace{\rho\frac{\partial \mathbf{u}}{\partial t}}_{\text{Unsteady}} + \underbrace{\rho(\mathbf{u} \cdot \nabla) \mathbf{u}}_{\text{Convection (Non-linear)}} = \underbrace{-\nabla p}_{\text{Pressure}} + \underbrace{\mu \nabla^{2} \mathbf{u}}_{\text{Diffusion}} + \mathbf{f}
    \end{align}
    \item \textbf{Scalar Transport Equation (Convection-Diffusion):} Describes how a passive scalar quantity $c$ (e.g., concentration, heat) evolves within a given velocity field $\mathbf{u}$:
    \begin{align}
        \frac{\partial c}{\partial t} + \underbrace{\nabla \cdot (\mathbf{u}c)}_{\text{Convection}} = \underbrace{D\nabla^{2}c}_{\text{Diffusion}} + S \label{eq:transport}
    \end{align}
\end{itemize}

\textbf{Theoretical Selection:} Directly mapping the Momentum Equation to GNNs is theoretically ill-posed and computationally prohibitive.
\begin{itemize}
    \item \textbf{Nature of Features:} Node features $\mathbf{H}$ in GNNs represent abstract information states (analogous to concentration or heat profile) rather than kinematic momentum. They do not possess "inertia" in the Newtonian sense.
    \item \textbf{Complexity:} The momentum equation involves the non-linear advection term $(\mathbf{u} \cdot \nabla)\mathbf{u}$, which implies that velocity transports itself. Modeling this self-advection on graphs introduces unnecessary high-order complexity.
\end{itemize}

Consequently, we identify the Scalar Transport Equation (Eq. \ref{eq:transport}) as the correct isomorphism for Graph Neural Networks. In this analogy, the evolution of node features $\mathbf{H}$ corresponds to the transport of scalar $c$, driven by two competing mechanisms: \textbf{Diffusion} (smoothing via the Laplacian) and \textbf{Convection} (directed transport via the velocity field).

\textbf{Structural Mapping to Graphs:} To ground Eq. \ref{eq:transport} in graph topology, we perform the following discretizations:
\begin{itemize}
    \item \textbf{Diffusion Term ($D\nabla^{2}c \to \mathcal{L}\mathbf{H}$):} The Laplacian operator $\nabla^2$ naturally maps to the normalized graph Laplacian $\mathbf{L}$ or the diffusion matrix $\mathbf{\tilde{A}}$. This formulation underpins classical methods like GCN \cite{gcn} and GRAND \cite{grand}, which are essentially pure diffusion processes ($c_t = \nabla^2 c$).
    \item \textbf{Convection Term ($\nabla \cdot (\mathbf{u}c) \to \text{Proposed}$):} Conventional GNNs largely overlook this term. In GNSN, we reintroduce this mechanism. As derived in the previous sections, since the vector field $\mathbf{u}$ cannot be explicitly defined on a graph metric space, we implement convection and topology-constrained message passing.
\end{itemize}

By integrating these components, GNSN effectively solves the discrete graph counterpart of the Convection-Diffusion equation, enabling the system to transcend the over-smoothing limitations of pure diffusion models.

\subsection{Dirichlet Energy in GNSN}

To theoretically validate GNSN's capability to alleviate oversmoothing, we analyze the asymptotic behavior of node features through the lens of Dirichlet energy. The Dirichlet energy $E(\mathbf{H})$ quantifies the "roughness" or variability of the signal on the graph. Formally, it is defined as:
\begin{align}
    E(\mathbf{H}) = \frac{1}{N} \sum_{i \in \mathcal{V}} \sum_{j \in \mathcal{N}(i)} \mathbf{A}_{ij} \|\mathbf{h}_{i}-\mathbf{h}_{j}\|^{2} = \frac{1}{N} \mathrm{Tr}\left(\mathbf{H}^\top \widetilde{\mathbf{L}}\,\mathbf{H}\right),
\end{align}
where $\widetilde{\mathbf{L}}$ is the graph Laplacian. The phenomenon of \textit{oversmoothing} corresponds to the exponential decay of this energy, i.e., $E(t) \to 0$ as $t \to \infty$, implying that all node features converge to a uniform, indistinguishable state (low-frequency limit).

\textbf{Energy Evolution Dynamics} \\
The time derivative of the energy characterizes the feature evolution trajectory:
\begin{align}
    \frac{dE}{dt} = \frac{1}{N} \mathrm{Tr}\left((\partial_t\mathbf{H})^\top \widetilde{\mathbf{L}}\,\mathbf{H} + \mathbf{H}^\top \widetilde{\mathbf{L}}\,(\partial_t\mathbf{H})\right) = \frac{2}{N} \mathrm{Tr}\left((\partial_t\mathbf{H})^\top \widetilde{\mathbf{L}}\,\mathbf{H}\right).
\end{align}
Substituting the continuous-time governing equation of GNSN (where the total flow is a weighted combination of convection and diffusion):
\begin{equation}
    \partial_t\mathbf{H} = \underbrace{(1 - H^*_\mathcal{G}) (\mathbf{u}_\mathcal{G} \odot \mathbf{\tilde{A}}\mathbf{H})}_{\text{Convection}} - \underbrace{H^*_\mathcal{G} (\widetilde{\mathbf{L}}\mathbf{H})}_{\text{Diffusion}},
\end{equation}
we derive the energy evolution equation:
\begin{align}
    \frac{dE}{dt} &= \frac{2}{N} \left[ (1 - H^*_\mathcal{G}) \mathrm{Tr}\left( (\mathbf{u}_\mathcal{G} \odot \mathbf{\tilde{A}}\mathbf{H})^\top \widetilde{\mathbf{L}}\mathbf{H} \right) - H^*_\mathcal{G} \mathrm{Tr}\left( (\widetilde{\mathbf{L}}\mathbf{H})^\top \widetilde{\mathbf{L}}\mathbf{H} \right) \right] \nonumber \\
    &= \frac{2}{N} \left[ \underbrace{(1 - H^*_\mathcal{G}) \sum_{i} \mathbf{u}_i (\mathbf{\tilde{A}}\mathbf{H})_i^\top (\widetilde{\mathbf{L}}\mathbf{H})_i}_{\mathcal{R}_{\text{conv}}: \text{Convective Source}} - \underbrace{H^*_\mathcal{G} \|\widetilde{\mathbf{L}}\mathbf{H}\|_F^2}_{\mathcal{R}_{\text{diff}}: \text{Diffusive Decay}} \right].
\end{align}

We analyze the distinct roles of the two components:

\textbf{Pure Diffusion Regime ($H^*_\mathcal{G} \to 1$):} The equation simplifies to $\frac{dE}{dt} \propto -\|\widetilde{\mathbf{L}}\mathbf{H}\|_F^2 \le 0$.
Here, the derivative is strictly non-positive. The system follows the gradient descent of the Dirichlet energy, forcing the graph signal to become globally smooth. This mathematically guarantees the occurrence of oversmoothing in deep diffusion-based GNNs.

\textbf{Pure Convection Regime ($H^*_\mathcal{G} \to 0$):}
The evolution is governed by $\mathcal{R}_{\text{conv}}$. Unlike the diffusive term, the sign of $\sum \mathbf{u}_i (\mathbf{\tilde{A}}\mathbf{H})_i^\top (\widetilde{\mathbf{L}}\mathbf{H})_i$ is \textbf{indefinite}. It depends on the alignment between the incoming convective flux $(\mathbf{\tilde{A}}\mathbf{H})_i$ and the local Laplacian curvature $(\widetilde{\mathbf{L}}\mathbf{H})_i$. 
Crucially, the scalar intensity $\mathbf{u}_i$ acts as an energy injection gain. By learning appropriate $\mathbf{u}_i$, the model can amplify high-frequency signals, effectively enabling $\frac{dE}{dt} > 0$ locally or globally. This "active transport" prevents the features from collapsing into a trivial equilibrium.

\textbf{GNSN Equilibrium ($0 < H^*_\mathcal{G} < 1$):}
GNSN operates as a dynamic competition between smoothing and transport:
\begin{align}
    \frac{dE}{dt} = \mathcal{R}_{\text{conv}} - \mathcal{R}_{\text{diff}}.
\end{align}
While diffusion ($\mathcal{R}_{\text{diff}}$) continuously dissipates high-frequency information to enforce homophily, convection ($\mathcal{R}_{\text{conv}}$) re-injects structural information based on transport intensity $\mathbf{u}_i$. This balance allows GNSN to maintain distinguishable node representations (non-zero energy) even at infinite depth, theoretically resolving the oversmoothing paradox.

\subsection{Parameter Efficiency and Numerical Stability}

\textbf{Implicit Parameterization:} Consistent with the finite-volume formulation in Section 3.2, the node-specific velocity $\mathbf{u}_i$ is not introduced as a free learnable parameter (which would increase model complexity). Instead, it is strictly derived from the accumulated feature flux via the inverse inference mechanism defined in Eq.~\ref{eq:eqvelo}. 
At each integration step, the aggregated incoming messages quantify the instantaneous transport intensity. The final velocity magnitude is obtained by integrating this intensity along the ODE trajectory. This design ensures that $\mathbf{u}_i$ is structure-aware—dynamically reflecting the actual information throughput determined by the graph topology and feature evolution—without adding extraneous trainable weights.

\textbf{Velocity Saturation Mechanism:} In the context of Neural ODEs, unbounded derivative terms can lead to numerical stiffness and instability. To prevent exploding convection terms, we impose a physical saturation constraint on the transport intensity. Let $\hat{\mathbf{u}}_i$ denote the raw computed integral from Eq.~\ref{eq:eqvelo}. The effective velocity used for state updates is bounded via a saturation (clipping) operation:
\begin{align}
    \mathbf{u}_i = \min\big(\hat{\mathbf{u}}_i, \mathbf{u}_{\max}\big),
\end{align}
where $\mathbf{u}_{\max}$ is a predefined hyperparameter acting as a "speed limit." This constraint prevents excessively large transport intensities that could destabilize the solver, particularly in early training stages or on graphs with high-degree hubs.

\textbf{Boundedness via Normalization:} Furthermore, the interaction weights $\alpha_{ij}$ are normalized (e.g., via Softmax), satisfying $\sum_{j \in \mathcal{N}(i)} \alpha_{ij} = 1$. This ensures that the instantaneous flux aggregation behaves as a convex combination (a non-expansive operator) with respect to the neighbor features. 
The combination of normalized flux and saturated velocity guarantees that the convection dynamics remain Lipschitz continuous and numerically bounded. This provides strong theoretical assurance for stable training, even when integrating over long temporal horizons (deep layers).

\newpage
\section{Visualization}
\label{appendixE}
\begin{figure*}[h]
    \centering

    \begin{subfigure}[t]{0.3\textwidth}
    \centering
        \includegraphics[width=0.9\linewidth]{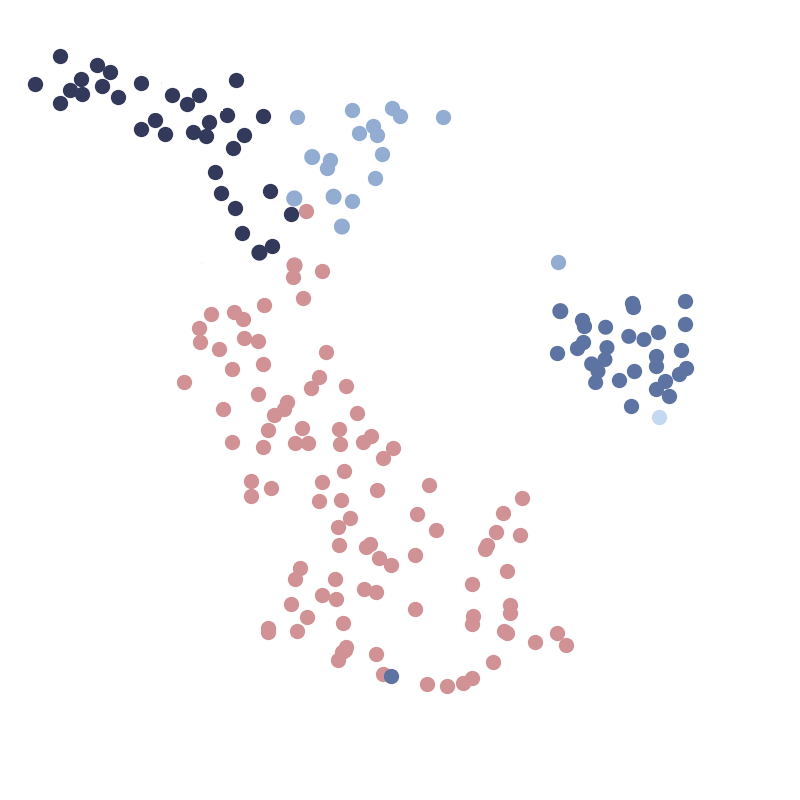}
        \caption{Texas}
    \end{subfigure}
    \hfill
    \begin{subfigure}[t]{0.3\textwidth}
    \centering
        \includegraphics[width=0.9\linewidth]{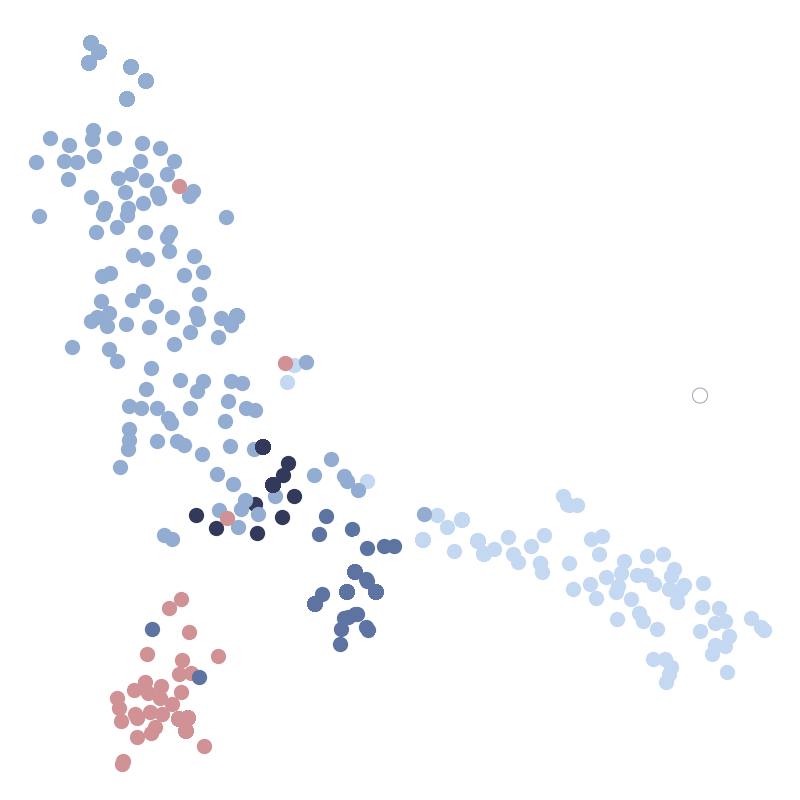}
        \caption{Wisconsin}
    \end{subfigure}
    \hfill
    \begin{subfigure}[t]{0.3\textwidth}
    \centering
        \includegraphics[width=0.9\linewidth]{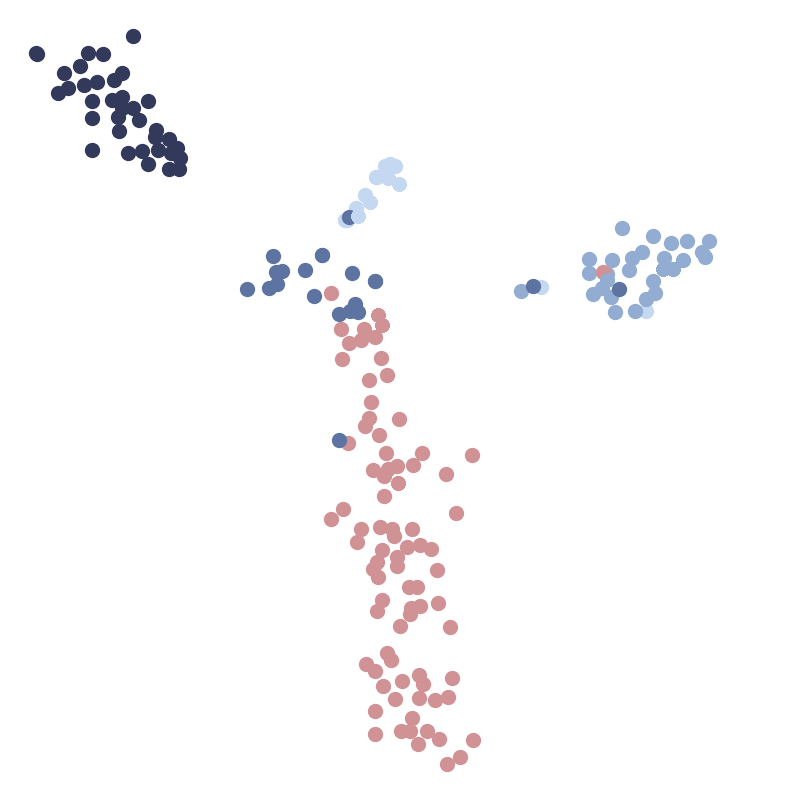}
        \caption{Cornell}
    \end{subfigure}

    \medskip

    \begin{subfigure}[t]{0.3\textwidth}
    \centering
        \includegraphics[width=0.9\linewidth]{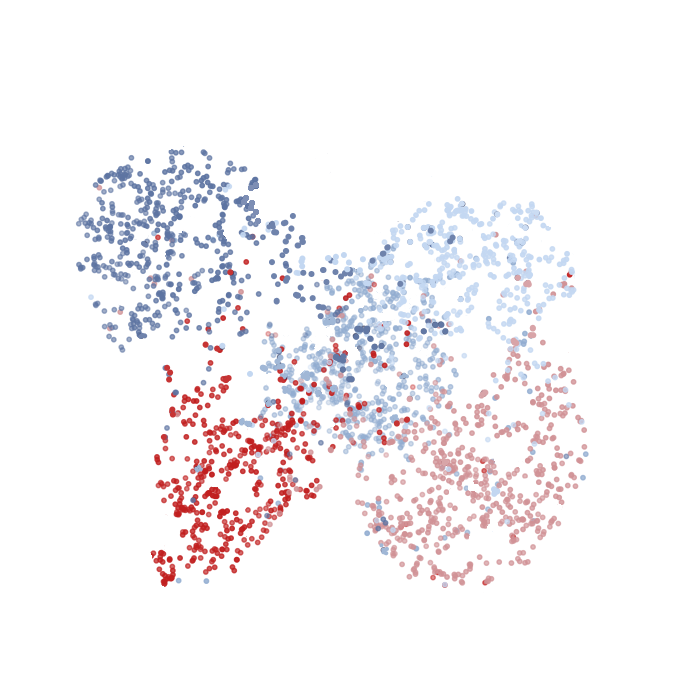}
        \caption{Chameleon}
    \end{subfigure}
    \hfill
    \begin{subfigure}[t]{0.3\textwidth}
    \centering
        \includegraphics[width=0.9\linewidth]{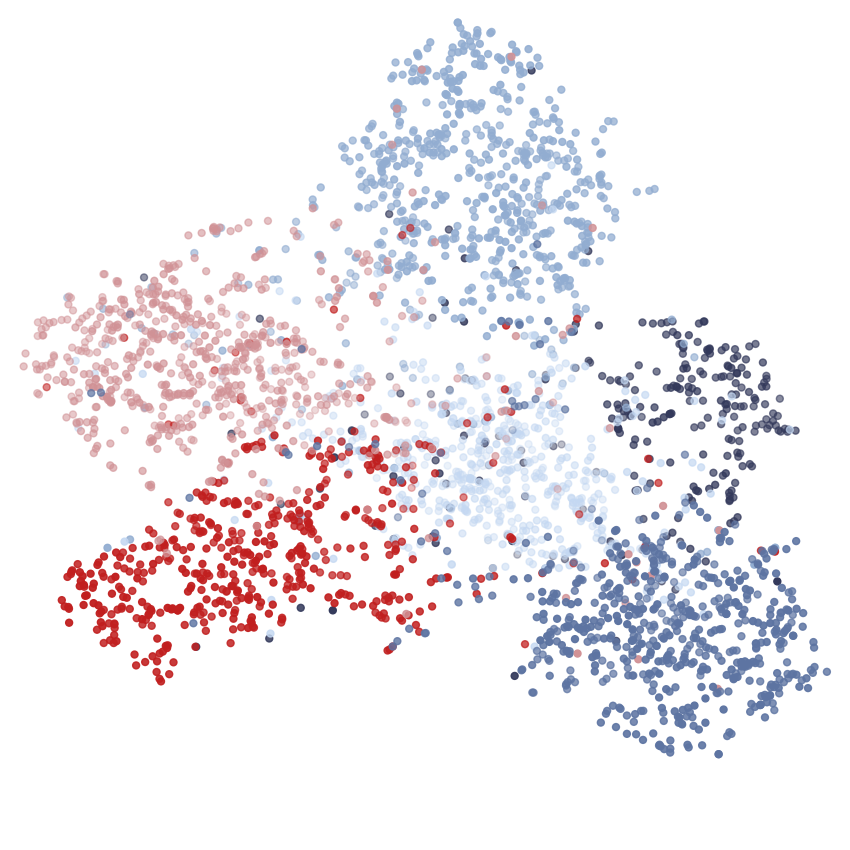}
        \caption{Citeseer}
    \end{subfigure}
    \hfill
    \begin{subfigure}[t]{0.3\textwidth}
    \centering
        \includegraphics[width=0.9\linewidth]{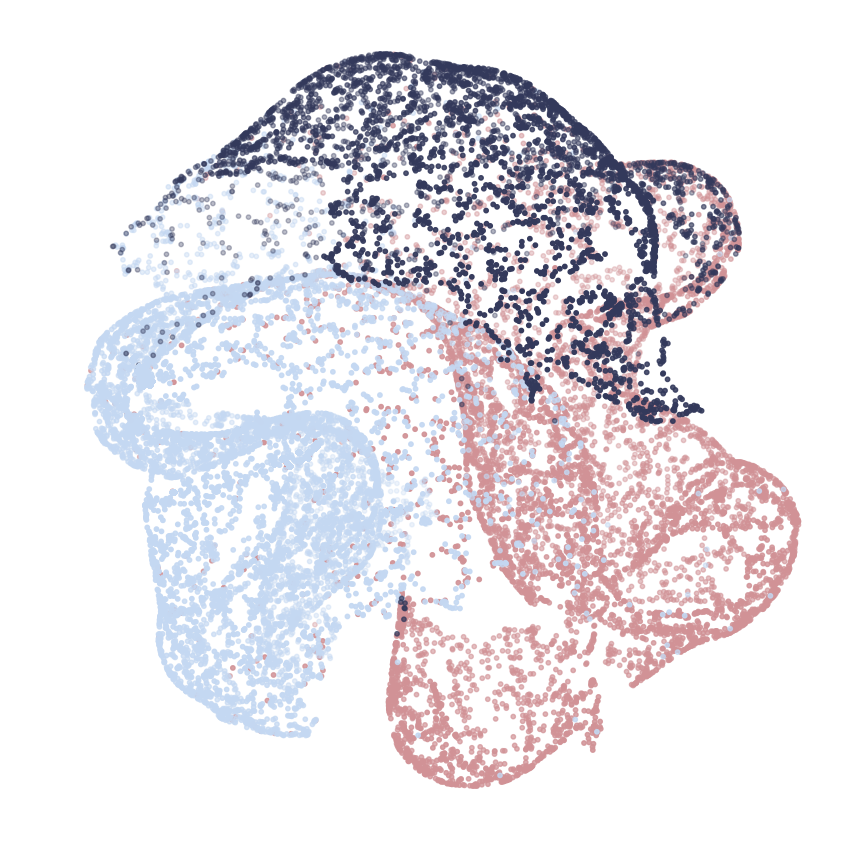}
        \caption{PubMed}
    \end{subfigure}

    \medskip

    \begin{subfigure}[t]{0.3\textwidth}
    \centering
        \includegraphics[width=0.9\linewidth]{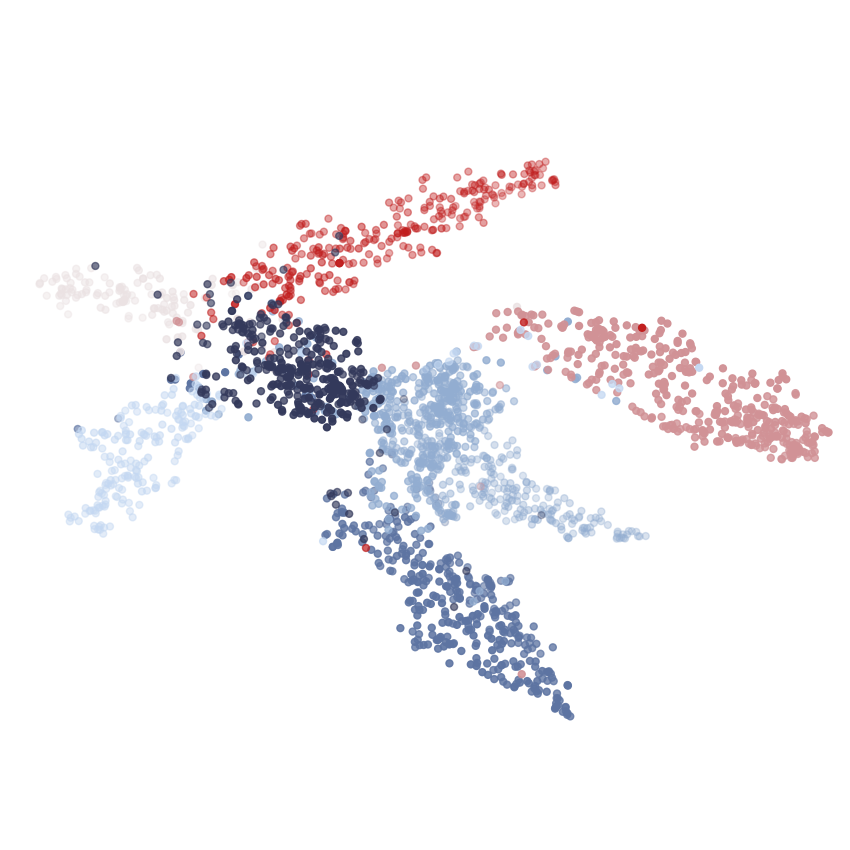}
        \caption{Cora}
    \end{subfigure}
    \hfill
    \begin{subfigure}[t]{0.3\textwidth}
    \centering
        \includegraphics[width=0.9\linewidth]{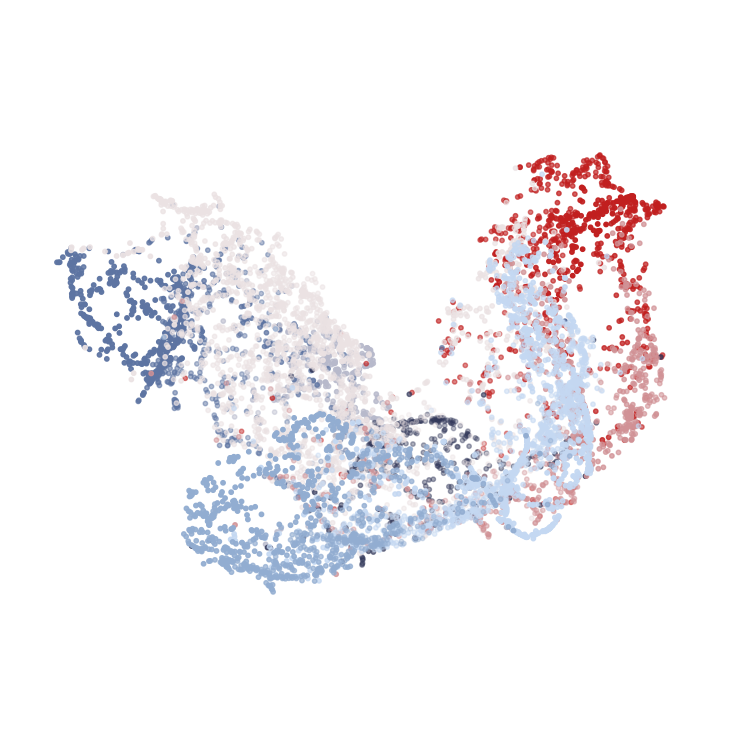}
        \caption{Photo}
    \end{subfigure}
    \Description{Visualization results on eight real-world datasets.}
    \caption{Visualization results on eight real-world datasets.}
    \label{fig:fig7}
\end{figure*}

In order to intuitively demonstrate and validate the effectiveness and superior performance of our proposed model, we further conduct visualization tasks on eight real-world datasets (Texas, Wisconsin, Chameleon, Cornell, Citeseer, PubMed, Cora and Photo). We extract the output vectors from the final layer of GNSN, project the node representations into a lower-dimensional space, and utilize t-SNE for visualization. Figure \ref{fig:fig7} presents the visualization results across each dataset, where different colors represent distinct ground-truth classes.

\newpage
\section{Sensitivity Analyses}
\label{appendixF}
\subsection{Sensitivity w.r.t. the ODE solver}

In GNSN, the continuous-time dynamic message passing mechanism is implemented by Neural ODEs. Previous studies in this domain have predominantly employed the following three ODE solvers \cite{grand,acmp,gread}:

\textbf{Euler Method (Euler):} The Euler method represents the most fundamental numerical integration technique, characterized by a first-order local truncation error of \(O(h)\). While computationally efficient, it suffers from significant error accumulation \cite{euler}.

\textbf{Fourth-Order Runge-Kutta Method (RK4):} The RK4 method offers a substantial improvement in accuracy, achieving a local truncation error of 
\(O(h^{4})\). By utilizing intermediate evaluations of the function derivatives, it provides a balance between computational cost and numerical stability \cite{rk4}.

\textbf{Dormand-Prince Fifth-Order Method (Dopri5):} The Dopri5 method is an adaptive-step solver that employs an embedded fourth- and fifth-order Runge-Kutta scheme to dynamically adjust the step size based on local error estimates. With an error order of \(O(h^{5})\) \cite{dopri5}.

\begin{figure*}[h]
    \centering
    \begin{minipage}{0.49\textwidth}
        \centering
        \includegraphics[width=\textwidth]{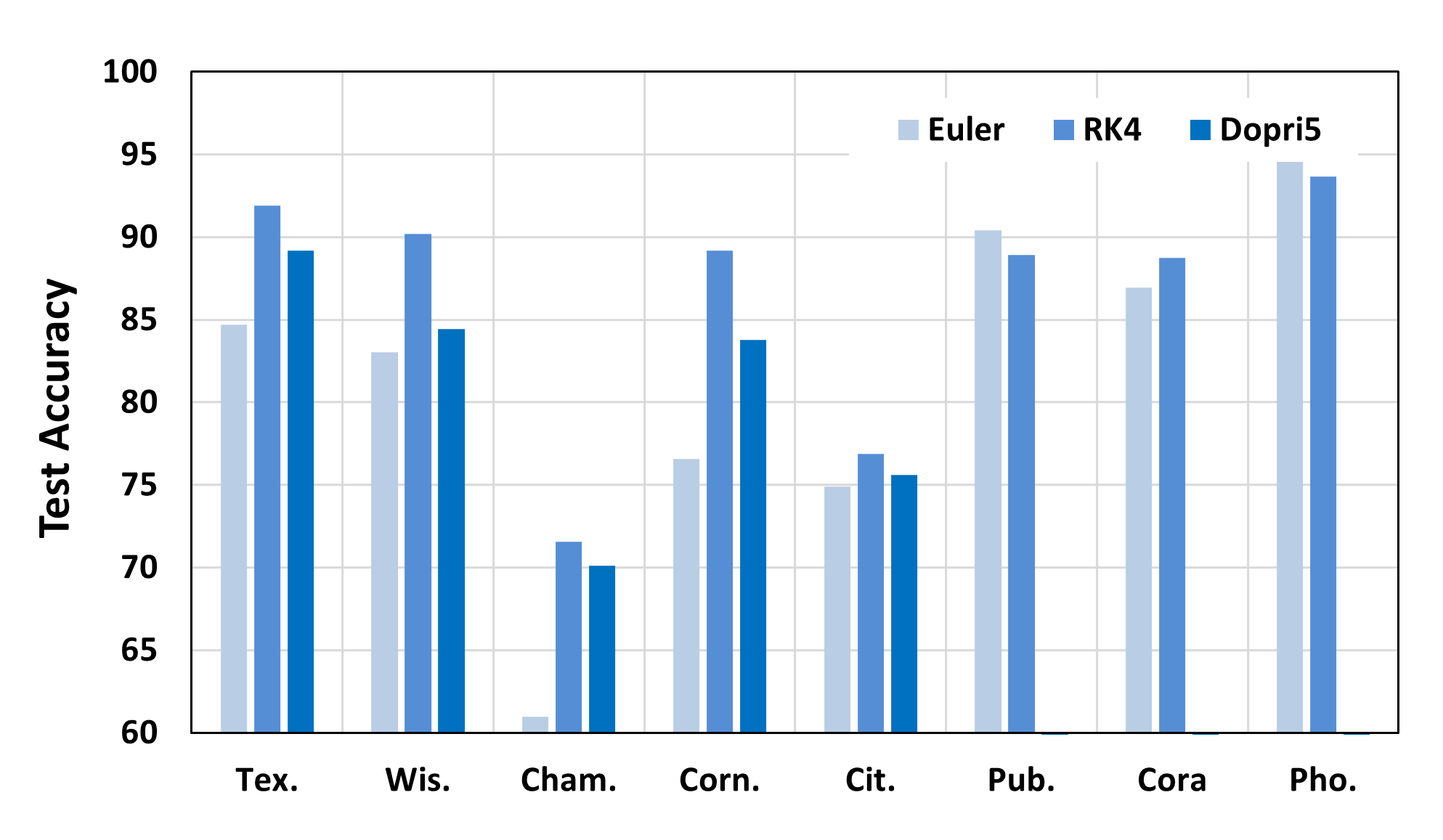} 
        \Description{}
        \caption{Sensitivity analysis}
        \label{fig:fig8}
    \end{minipage}%
    \begin{minipage}{0.43\textwidth}
        \centering
        \includegraphics[width=\textwidth]{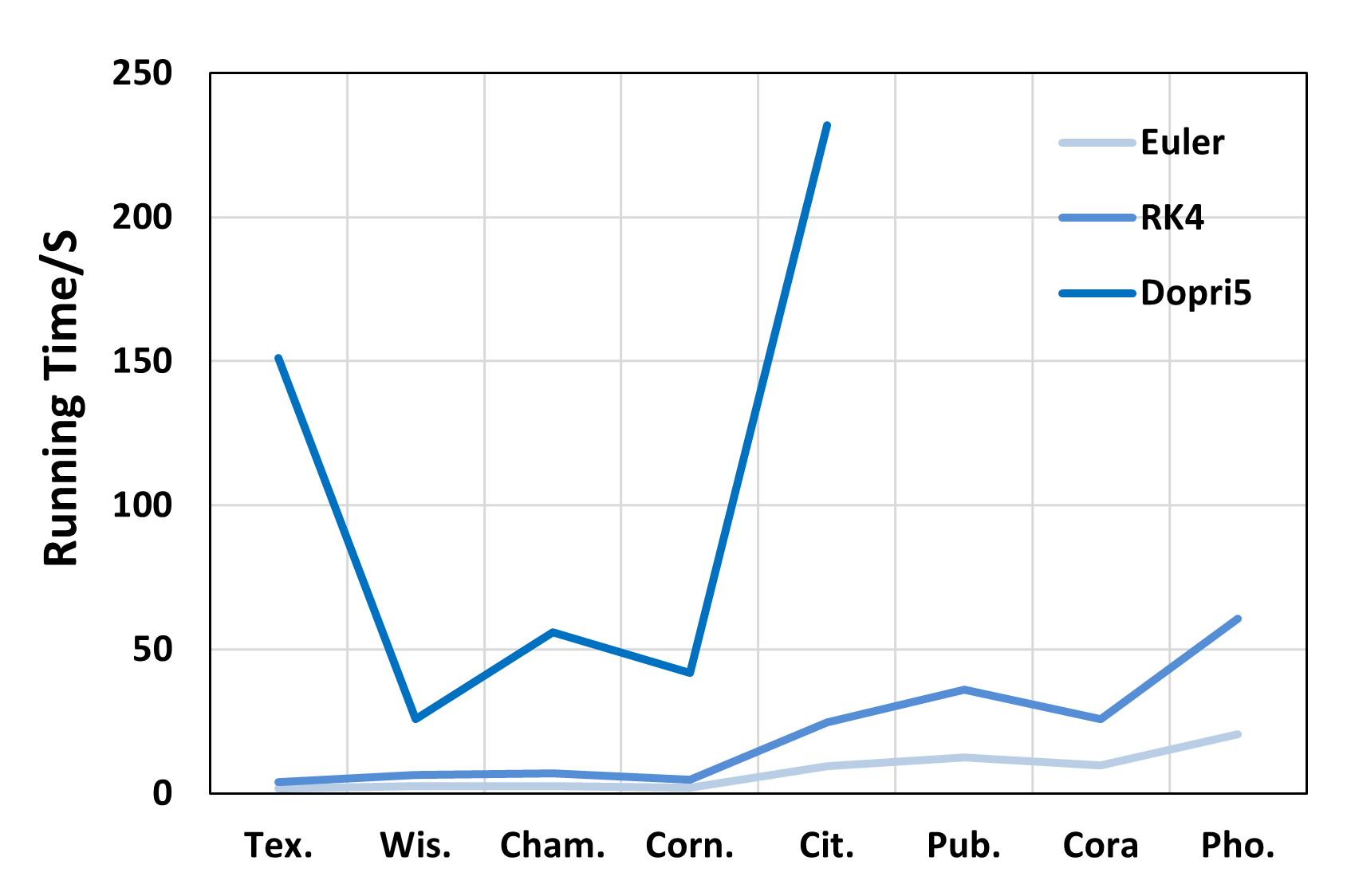}
        \Description{}
        \caption{Training time consumption}
        \label{fig:fig9}
    \end{minipage}%
\end{figure*}

In Table.\ref{tab:tab11} and Figure \ref{fig:fig8}, we present the performance of three ODE solvers across eight real-world datasets. Due to memory constraints, the Dopri5 solver fails to complete computation on Pubmed, Cora and Photo, both of which contain a large number of nodes and edges. Figure \ref{fig:fig9} illustrates the total training time consumption of for 200 epochs across different datasets for each ODE solver. From the above results, RK4 demonstrates strong adaptability to GNSN. Except for the Pubmed and Photo dataset, RK4 consistently achieves the best performance while maintaining a relatively low time consumption, making it a practical choice for balancing accuracy and efficiency in various scale datasets.
\begin{table*}[ht]
\centering
\caption{Sensitivity analysis of ODE solvers on eight real-world datasets.}
\begin{tabular}{ccccccccc}
\toprule
\textbf{Dataset} & \textbf{Texas} & \textbf{Wisconsin} & \textbf{Chameleon} & \textbf{Cornell} & \textbf{Citeseer} & \textbf{PubMed} & \textbf{Cora} &\textbf{Photo} \\
\midrule
Euler   & 84.68 & 83.01   &60.99  & 76.58   & 74.92    & \textcolor{red}{90.41}  & 86.92 & \textcolor{red}{95.43}\\
RK4     & \textcolor{red}{91.89} & \textcolor{red}{90.20}   & \textcolor{red}{71.58}  & \textcolor{red}{89.19}   & \textcolor{red}{76.87}    & 88.93  & \textcolor{red}{88.73} & 93.66 \\
Dopri5  & 89.19 & 84.42  & 70.11   & 83.78   & 75.61    & N/A    & N/A  & N/A \\ \bottomrule

\end{tabular}

\label{tab:tab11}
\end{table*}

\newpage
\subsection{Sensitivity w.r.t. the ODE step size}
\begin{figure}[ht]
\begin{center}
\centerline{\includegraphics[width=0.5\columnwidth]{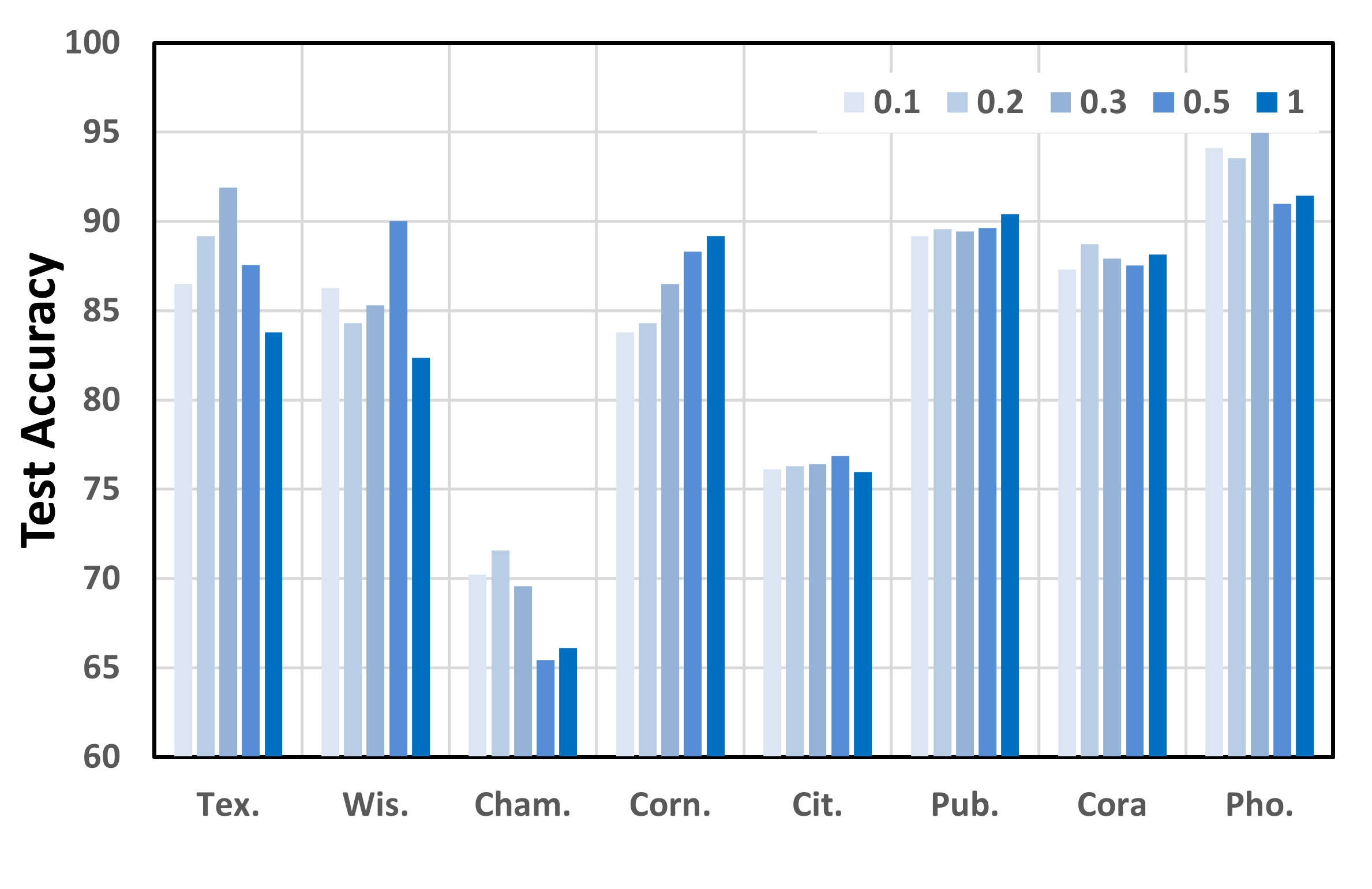}}
\Description{}
\caption{Sensitivity analysis of the ODE step size}
\label{fig:fig10}
\end{center}
\end{figure}

Here, we investigate the impact of ODE solver step size on the performance of GNSN across eight real-world datasets. Specifically, Euler is used for Pubmed and Photo, while RK4 is applied to the remaining datasets. From Table \ref{tab:tab12} and Figure \ref{fig:fig10}, we observe that for heterophilic datasets, variations in step size lead to significant fluctuations in test accuracy, indicating a strong sensitivity to this hyperparameter. In contrast, for homophilic datasets, test accuracy remains relatively stable across different step sizes. However, no clear pattern emerges regarding the optimal step size preference across different datasets.

\begin{table*}[ht]
\centering
\caption{Sensitivity analysis of the ODE step size.}
\begin{tabular}{ccccccccc}
\toprule
Step Size & Texas & Wisconsin & Chameleon & Cornell & Citeseer & Pubmed & Cora & Photo \\
\midrule
0.1 & 86.49 & 86.27 & 70.22 & 83.78 & 76.13 & 89.17 & 87.32 & 94.11 \\
0.2 & 89.19 & 84.31 & \textcolor{red}{71.58} & 84.31 & 76.28 & 89.58 & \textcolor{red}{88.73} & 93.55 \\
0.3 & \textcolor{red}{91.89} & 85.29 & 69.57 & 86.49 & \textcolor{red}{76.43} & 89.43 & 87.93 & \textcolor{red}{95.43} \\
0.5 & 87.56 & \textcolor{red}{90.02} & 65.44 & 88.31 & 76.87 & 89.63 & 87.53 & 91.00 \\
1.0 & 83.78 & 82.35 & 66.12 & \textcolor{red}{89.19} & 75.98 & \textcolor{red}{90.41} & 88.15 & 91.44 \\
\bottomrule
\end{tabular}
\label{tab:tab12}
\end{table*}

\newpage

\subsection{Learnable homophily ratio}

The homophily ratio \( H_{\mathcal{G} }\) quantifies the degree to which adjacent nodes in a graph share the same class label, serving as a crucial global indicator of the alignment between the graph structure and label distribution. However, because its computation requires access to labels across the entire graph, directly incorporating 
\( H_{\mathcal{G} }\) into the model may lead to information leakage from the validation or test sets.

To address this issue, we introduce a learnable homophily ratio 
\(H^{\ast }_\mathcal{G}\), which is dynamically estimated using only the training data. As a trainable parameter, \(H^{\ast }_\mathcal{G}\) is continuously updated during training and is utilized to modulate the relative intensities of diffusion and convection flows within the message-passing process. Compared to the static prior \( H_{\mathcal{G} }\), the data-driven and dynamic nature of 
\(H^{\ast }_\mathcal{G}\)
improves the model’s generalization ability across graphs with varying degrees of homophily.

We define the graph homophily ratio estimated on the training set, denoted as \( H^*_\mathcal{G} \), as follows:
\begin{equation}
H^*_\mathcal{G} = \frac{1}{|\mathcal{V}_{train}|} \sum_{i \in \mathcal{V}_{train}} h_{\mathcal{N}1}(i) = \frac{1}{|\mathcal{V}_{train}|} \sum_{i\in \mathcal{V}_{train}} \frac{|{j \in \mathcal{N}_1(i) : y_j = y_i}|}{|\mathcal{N}_1(i)|}
\end{equation}
where \( \mathcal{V}_{train} \) denotes the set of nodes in training set, \( y_i \) and \( y_j \) are the ground-truth class labels of nodes \( i \) and \( j \). Since the indicator function is non-differentiable and cannot be used in gradient-based optimization, we propose a differentiable surrogate. During training, we replace one-hot labels with soft predictions (class probability vectors), and define a similarity function using cosine similarity:
\begin{equation}
\text{sim}(y_i, y_j) = \cos(y_i, y_j)
\end{equation}

We then apply the sigmoid function to normalize the aggregated similarity over the training edges, resulting in a differentiable version of the homophily ratio:
\begin{equation}
H^*_\mathcal{G} = \sigma\left( 
\frac{1}{|\mathcal{V}_{train}|} \sum_{i \in \mathcal{V}_{train}} 
\frac{1}{|\mathcal{N}_1(i) \cap \mathcal{V}_{train}|} 
\sum_{j \in \mathcal{N}_1(i) \cap \mathcal{V}_{train}} 
\text{sim}(y_i, y_j)
\right),
\end{equation} where $\sigma(\cdot)$ denotes the sigmoid function, used to ensure the output $H^*_\mathcal{G}$ remains within the range $(0, 1)$.

\begin{figure}[ht]
\begin{center}
\centerline{\includegraphics[width=0.50\columnwidth]{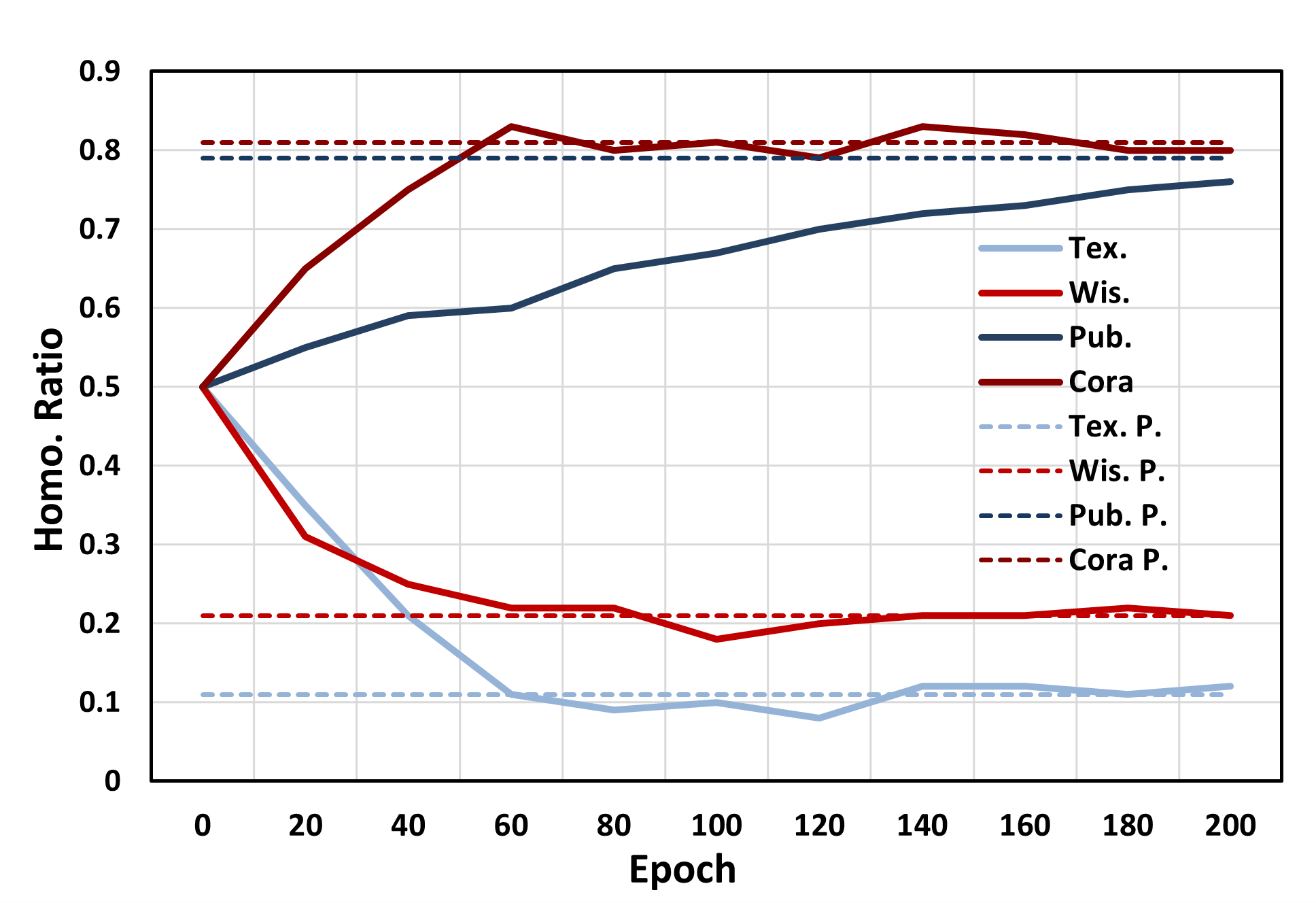}}
\Description{}
\caption{Training evolution of \(H^*_\mathcal{G}\).}
\label{fig:figa4}
\end{center}
\end{figure}

To validate the effectiveness of \(H^*_\mathcal{G}\), we compare its evolution over 200 training epochs with the prior homophily ratio \(H_\mathcal{G}\) (eg. Tex. P.) across different datasets—two heterophilic and two homophilic—for better visual clarity, as illustrated in the Figure \ref{fig:figa4}. To prevent information leakage and ensure fairness across datasets, we initialize \(H^*_\mathcal{G}\) to a uniform value of 0.5. It can be observed that as training progresses, 
\(H^*_\mathcal{G}\) gradually approaches the prior \(H_\mathcal{G}\). Although a certain gap remains, \(H^*_\mathcal{G}\)is able to reflect the label distribution characteristics of each dataset with reasonable accuracy.

To further investigate the impact of different formulations of the graph homophily ratio on model performance, we conduct a sensitivity analysis using three experimental settings: 

(1) \textbf{Prior-based group}: which uses the prior homophily ratio \(H_\mathcal{G}\) as the control factor for modulating the intensity of diffusion and convection.
\begin{align*}
    \frac{\partial \mathbf{H}(t)}{\partial t} + (1-H_\mathcal{G}) \mathbf{A}^\varepsilon \mathbf{H}(t)\cdot\nabla\mathbf{u}_\mathcal{G} =H_\mathcal{G}\mathbf{\widetilde{L} }\mathbf{H}(t)
\end{align*}

(2) \textbf{Random-initialized group}: where the modulation parameter is a randomly initialized learnable scalar without boundary constraints.
\begin{align*}
    \frac{\partial \mathbf{H}(t)}{\partial t} +  \beta\mathbf{A}^\varepsilon \mathbf{H}(t)\cdot\nabla\mathbf{u}_\mathcal{G} = \gamma\mathbf{\widetilde{L} }\mathbf{H}(t),
\end{align*} where \(\beta\) and \(\gamma\) are both learnable parameters. 

(3) \textbf{Adaptive group}: which employs the dynamically updated \(H^*_\mathcal{G}\) to adjust the intensity of diffusion and convection flows.
\begin{align*}
    \frac{\partial \mathbf{H}(t)}{\partial t} + (1-H^{\ast }_\mathcal{G}) \mathbf{A}^\varepsilon \mathbf{H}(t)\cdot\nabla\mathbf{u}_\mathcal{G} =H^{\ast }_\mathcal{G}\mathbf{\widetilde{L} }\mathbf{H}(t)
\end{align*}

\begin{figure}[ht]
\begin{center}
\centerline{\includegraphics[width=0.5\columnwidth]{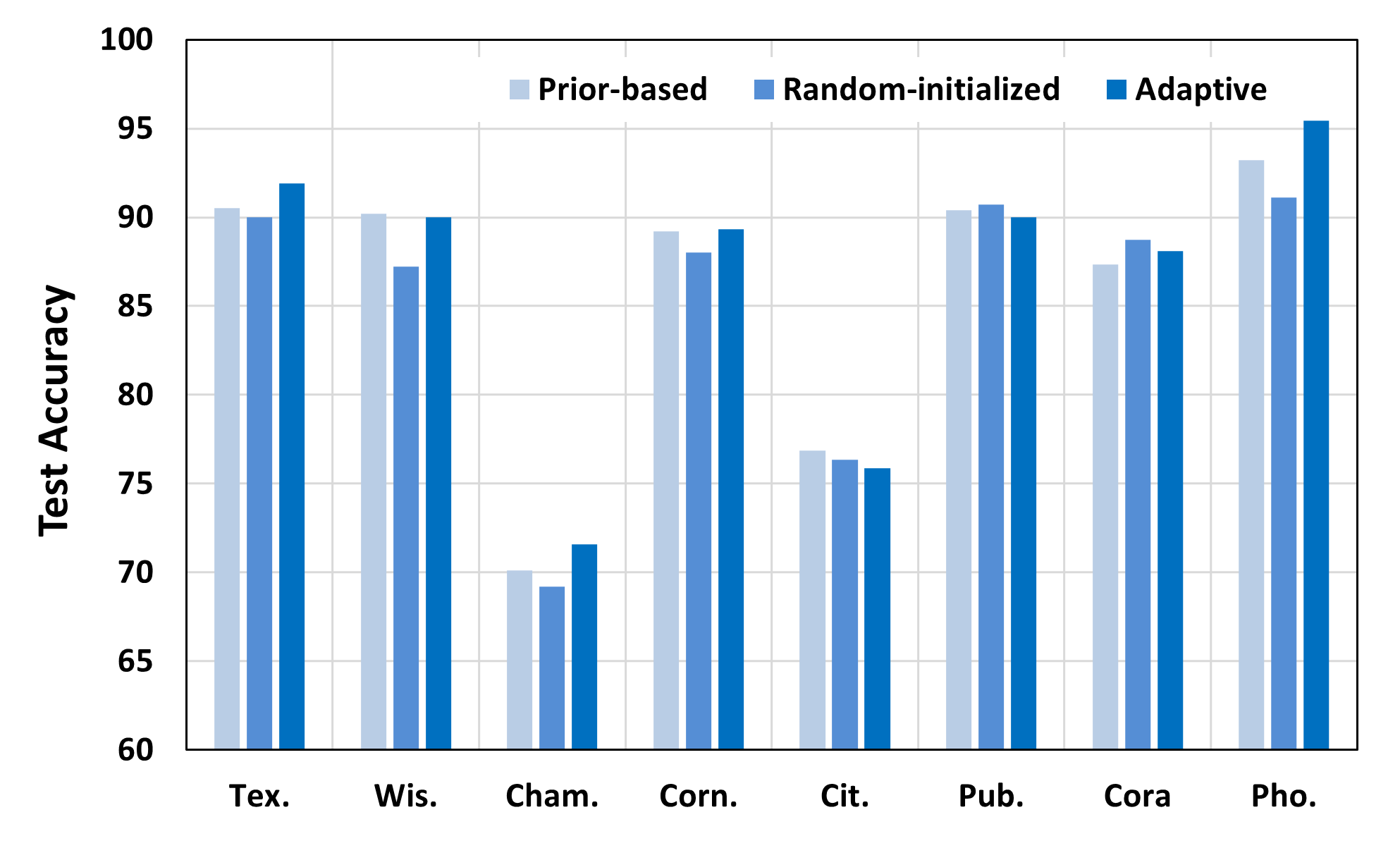}}
\Description{}
\caption{Sensitivity analysis of the formulations of intensity of diffusion and convection.}
\label{fig:figa5}
\end{center}
\end{figure}

As shown in Figure \ref{fig:figa5}, since \(H^*_\mathcal{G}\) is dynamicly updated during training and gradually approaches the prior\(H_\mathcal{G}\), the performance of the Prior-based group and the Adaptive group remains relatively close across most datasets. In contrast, the Random-initialized group exhibits inconsistent results—occasionally performing well, but lacking stability overall.

\newpage
\subsection{Position Encoding}
In the position encoding module, node features are projected from Euclidean space to the Poincaré disk model in hyperbolic space to serve as model inputs. To further evaluate the effectiveness of hyperbolic positional encoding, we perform a sensitivity analysis under two alternative experimental configurations:

(1) \textbf{ Lorentz / Hyperboloid Model:} To map Euclidean features \(\mathbf{X} \in \mathbb{R}^{n \times d}\) onto the Lorentzian hyperboloid \(\mathbb{H}^d \subset \mathbb{R}^{d+1}\), we define the projection as:
\begin{align}
    \mathbf{X}_L = \left[ \sqrt{1 + \|\mathbf{X}\|^2} \,\middle\|\, \mathbf{X} \right] \in \mathbb{R}^{n \times (d+1)} \label{eq:lorentz_proj}
\end{align}
This projection guarantees that the Lorentzian inner product satisfies:
\begin{align}
    \langle \mathbf{X}_L, \mathbf{X}_L \rangle_{\mathcal{L}} = -1 \label{eq:lorentz_inner}
\end{align}

(2) \textbf{Tangent Space:} To map node features \(\mathbf{X}\) onto the tangent space at the origin point \(o = [1, \mathbf{0}] \in \mathbb{R}^{d+1}\), denoted as \(T_o \mathbb{H}^d\), we apply the logarithmic map:

\begin{align}
    \log_o(\mathbf{X}) = \frac{\operatorname{arcosh}(-\langle o, \mathbf{X} \rangle_{\mathcal{L}})}{\sqrt{\langle o, \mathbf{X} \rangle_{\mathcal{L}}^2 - 1}} \cdot \left( \mathbf{X} + \langle o, \mathbf{X} \rangle_{\mathcal{L}} \cdot o \right) \label{eq:logmap}
\end{align}

As a result, we obtain a representation \(\mathbf{X}_{\log} \in \mathbb{R}^{n \times d}\).

\begin{figure}[ht]
\begin{center}
\centerline{\includegraphics[width=0.50\columnwidth]{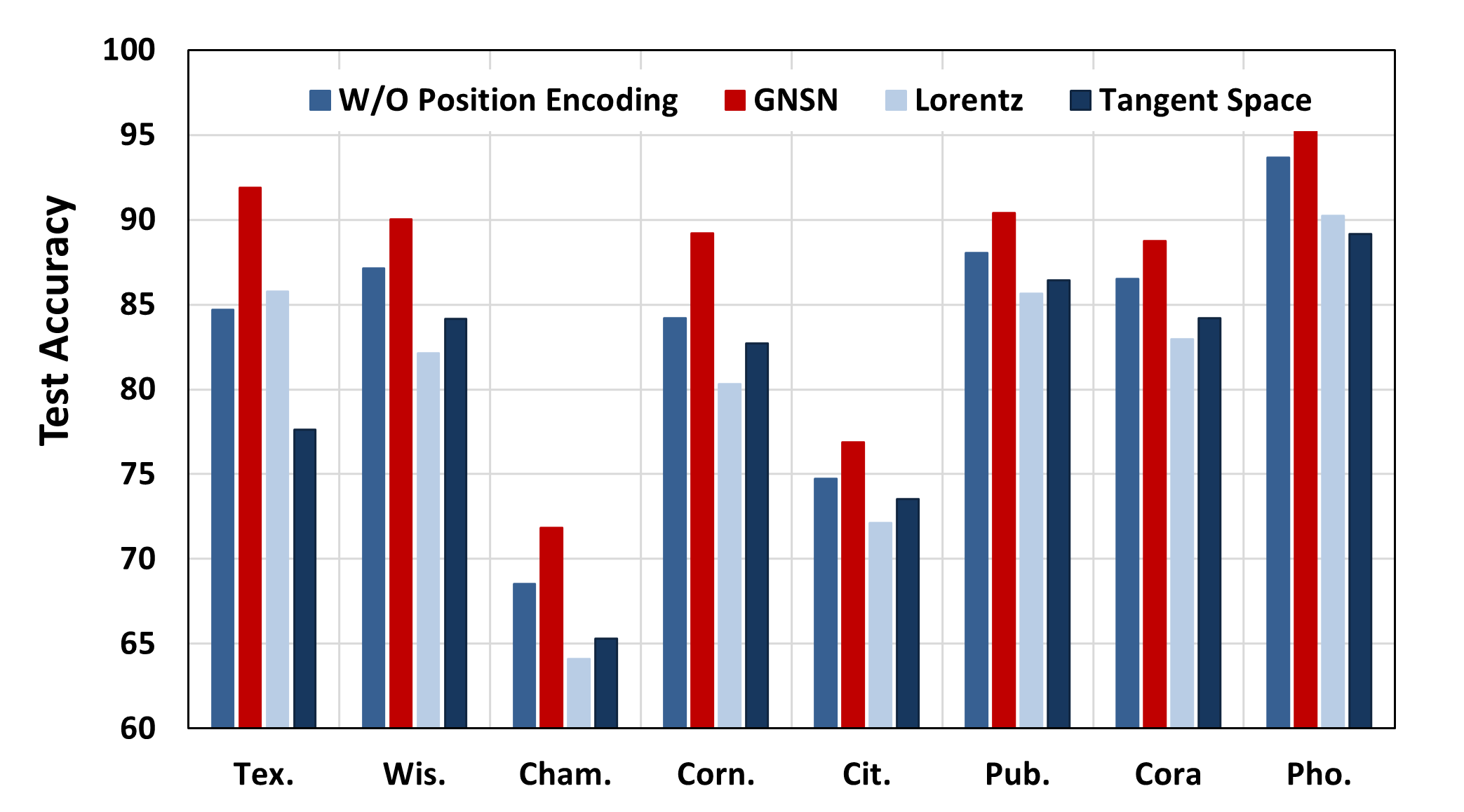}}
\Description{}
\caption{Sensitivity analysis of the hyperbolic encoding.}
\label{fig:figa6}
\end{center}
\end{figure}

As shown in Figure \ref{fig:figa6}, the another two hyperbolic encoding methods (Lorentz and Tangent Space) generally underperform compared to the GNSN (Poincaré disk). The Lorentz model shows unstable results across datasets, and the Tangent Space approach often leads to decreased accuracy, suggesting potential information loss during the projection.

\newpage
\section{Computational Complexity}
\label{appendixH}
\subsection{Complexity Analysis}
The computational complexity of GNSN primarily consists of two components: the diffusion message flow and the convection message flow.

\textbf{Diffusion complexity:} The diffusion term is dominated by the operation involving the \(\tilde{\mathbf{A} } \) , which has \(O(\left | \mathcal{E}  \right | )\) nonzero elements in a sparse setting. Multiplying this \(n\times n\) sparse matrix with a dense \(n\times d\) feature matrix results in a computational complexity of: \(O(\left | \mathcal{E}  \right |\cdot d )\)

\textbf{Convection Complexity:}
For the convection term, if the adjacency matrix 
\(\mathbf{A}\) is also sparse with \(O(\left | \mathcal{E}  \right | )\) nonzero elements, computing 
\(\mathbf{A}^{\varepsilon}\) requires:
\(O(\left | \mathcal{E}  \right |\cdot d_{\mathrm{max}}\varepsilon )\), where \(d_{\mathrm{max}}\) is the maximum node degree and 
\(\varepsilon\) represents the order of adjacency expansion. Multiplying \(\mathbf{A}^{\varepsilon}\) with the node feature matrix \(\mathcal{H}\) introduces an additional: \(O(\left | \mathcal{E}  \right |\cdot d )\). Besides, computing the node-wise velocity field \(\mathbf{u}\) has a complexity of \(O(n)\), leading to a final term of: \(O(n\cdot d )\). Since \(O(n\cdot d )\) is generally smaller than 
\(O(\left | \mathcal{E}  \right |\cdot d_{\mathrm{max}}\varepsilon )\) in sparse graphs, the dominant complexity for the convection term is:
\(O(\left | \mathcal{E}  \right |\cdot d_{\mathrm{max}}\varepsilon +\left | \mathcal{E}  \right |\cdot d   )\)

\textbf{Overall Complexity of GNSN:} Since both diffusion and convection terms share 
\(O(\left | \mathcal{E}  \right |\cdot d )\), the overall complexity simplifies to:
\(O(\left | \mathcal{E}  \right |\cdot d_{\mathrm{max}}\varepsilon +\left | \mathcal{E}  \right |\cdot d )\)

\subsection{Computational Overhead of the Velocity Field}

To quantitatively assess the computational overhead introduced by the convection term, we conducted a comparison between GNSN (with velocity field) and w/o velocity (GNSN without velocity field) across datasets. All experiments were performed for 100 training epochs under identical settings.

\begin{table}[h!]
\centering
\caption{Runtime and GPU memory consumption comparison.}
\begin{tabular}{llccccccccc}
\hline
 &  & Tex. & Wis. & Cham. & Corn. & Cit. & Pub. & Cora & Pho. & OGB-arxiv \\
\hline
\multirow{2}{*}{Time (s)} 
 & w/o velocity & 11.8 & 13.4 & 14.6 & 12.8 & 24.7 & 65.5 & 71.8 & 78.6 & 1219 \\
 & GNSN         & 14.2 & 16.3 & 17.8 & 15.6 & 30.2 & 80.4 & 74.8 & 95.5 & 1863 \\
\hline
\multirow{2}{*}{Used GPU Memory (GB)} 
 & w/o velocity & 0.89 & 0.81 & 0.97 & 0.86 & 1.37 & 3.06 & 1.85 & 3.33 & 6.89 \\
 & GNSN         & 1.15 & 1.02 & 1.26 & 1.09 & 1.65 & 3.78 & 2.32 & 4.05 & 9.46 \\
\hline
\label{tab:tab13}
\end{tabular}
\end{table}

As shown in Table \ref{tab:tab13}, the inclusion of the velocity field results in only a moderate increase in wall-clock runtime and GPU memory consumption, typically within 15\%–20\% for most datasets. While the overhead becomes more pronounced on larger and denser graphs, it remains well within a reasonable range, especially when weighed against the substantial performance gains enabled by velocity-driven modeling. Notably, the observed variations in runtime and resource usage are significantly smaller than those induced by different choices of ODE solvers, indicating that the learnable velocity field contributes only a minor and manageable computational cost relative to the overall model complexity.

\end{document}